\documentclass[a4paper, 10pt, 5p, twocolum]{elsarticle}

\usepackage{hyperref}
\hypersetup{colorlinks,citecolor=blue}

\usepackage{amsmath}
\usepackage{bm}
\usepackage{amssymb}
\usepackage{graphicx}
\usepackage{subcaption}
\usepackage{color}
\usepackage[usenames,dvipsnames,svgnames,table]{xcolor}
\usepackage{tabularx}
\usepackage{multirow}
\usepackage{url}

\usepackage{enumitem}   
\usepackage{xspace}
\usepackage{siunitx}
\usepackage{lineno}
\usepackage[baseline]{euflag}

\input{settings/def_macros.tex}

\begin{document}

\begin{frontmatter}
\title{Weakly and Semi-Supervised Detection, Segmentation and Tracking of Table Grapes with Limited and Noisy Data\fnref{fn2}}

\author[1]{Thomas A. Ciarfuglia\fnref{fn1}\corref{cor1}}
\author[1]{Ionut M. Motoi\fnref{fn1}} 
\author[1]{Leonardo Saraceni\fnref{fn1}}
\author[1]{Mulham Fawakherji}
\author[2]{Alberto Sanfeliu}
\author[1]{Daniele Nardi}

\fntext[fn1]{The authors contributed equally to the work.}%
\fntext[fn2]{This work extends the one titled "Pseudo-label Generation for Agricultural Robotics Applications" presented at the 3rd International Workshop on
Agriculture-Vision, CVPR 2022, New Orleans.}
\cortext[cor1]{Corresponding author {\tt\footnotesize ciarfuglia@diag.uniroma1.it}}%

\affiliation[1]{organization={Department of Information, Management and Automation Engineering (DIAG), Sapienza University of Rome}, addressline={via Ariosto 25}, city={Rome}, postcode={00185},country={Italy}}
\affiliation[2]{organization={Institut de Robotica i Informatica Industrial, CSIC-UPC},
addressline={Calle Llorens i Artigas 4-6}, postcode={08028}, city={Barcelona}, country={Spain}}


\begin{abstract}
Detection, segmentation and tracking of fruits and vegetables are three fundamental tasks for precision agriculture, enabling robotic harvesting and yield estimation applications. However, modern algorithms are data hungry and it is not always possible to gather enough data to apply the best performing supervised approaches. Since data collection is an expensive and cumbersome task, the enabling technologies for using computer vision in agriculture are often out of reach for small businesses. Following previous work in this context \citep{Ciarfuglia2022pseudo}, where we proposed an initial weakly supervised solution to reduce the data needed to get state-of-the-art detection and segmentation in precision agriculture applications, here we improve that system and explore the problem of tracking fruits in orchards. We present the case of vineyards of table grapes in southern Lazio (Italy) since grapes are a difficult fruit to segment due to occlusion, color and general illumination conditions. We consider the case in which there is some initial labelled data that could work as source data (\eg wine grape data), but it is considerably different from the target data (\eg table grape data). To improve detection and segmentation on the target data, we propose to train the segmentation algorithm with a weak bounding box label, while for tracking we leverage 3D Structure from Motion algorithms to generate new labels from already labelled samples. Finally, the two systems are combined in a full semi-supervised approach. Comparisons with state-of-the-art supervised solutions show how our methods are able to train new models that achieve high performances with few labelled images and with very simple labelling. 
\end{abstract}
\end{frontmatter}

\section{Introduction} \label{sec:introduction}

Detection and tracking of fruits and vegetables are two fundamental tasks for precision agriculture, enabling robotic harvesting and yield estimation application. As with any other automation task, detection of vegetables benefits from controlled environments and well known field conditions.
The reduction of variability and uncertainty in fruit position, occlusions, variety, illumination, to cite a few aspects of the problem, have a huge impact on the successful implementation of a learning based detection system. 
For this reason, many detection based systems are designed with the aim of reducing the sources of variability. For example, \citep{Wang2013automated} and \citep{NuskeJFR2014automated} proposed detection systems for fruit counting and yield estimation using a direct illumination device to control ambient light. In both cases, the proposed systems need to be run at night to be functional. In \citep{PrettoRAM2021building}, an example of a straddle robotic platform is given, which is another common way to control the environmental light and remove camera intrinsics variabilities.  

While these approaches are viable, they are also difficult to adapt to different cultivations and require considerable economical and technical investment, which is often beyond the capacities of small and medium agricultural businesses, which are often family based. While the economic problems they face are generally the same as those of bigger companies (e.g. lack of manpower to harvest vegetables), they do not have the economic strength or knowledge to engineer the cultivation from the ground up for heavily automated processes. This means that having more flexible approaches that are more algorithmic and data oriented than hardware oriented would positively impact these businesses, allowing a range of possible applications of data driven precision agriculture.   

In this this respect, modern computer vision, mostly based on deep learning algorithms, has lowered the the initial investment needed to integrate advanced detection techniques for monitoring and managing the crops. However, as discussed by \citep{KoiralaCEA2019deep} in their survey on deep learning techniques applied to fruit counting, these algorithms are data hungry and gathering the correct and right amount of data is not always straightforward. 

One way to face data scarcity is the use of algorithmic techniques of semi-supervised, weakly-supervised and transfer learning, where additional information is added to the training process as an external training signal, or by leveraging what was learned in a different but related task. An example of these approaches in agriculture is given by \citep{BellocchioRAL2019weakly} where the authors propose an olive counting solution that is explicitly trained with weak labels and consistency losses. The auxiliary signal in this case is the labelling obtained by an external classifier that detects whether or not there are olives in the picture. This work is close to ours for the focus on working on data with minimal labelling. However, it is based on simple direct fruit counting, which can lead to huge errors in cases where self occlusion is typical. 
An example of transfer learning that can be used to reduce the amount of data needed for training is shown in \citep{Guldenring2021self-supervised}. In this study, the authors show how pre-training using contrastive learning as an unsupervised technique is able to improve the performances of detection algorithms compared to standard ImageNet pre-training. This result is a good starting point for training a deep network, but  some task specific training data is still needed.   

In this context, it is interesting to consider the challenges that detection and tracking algorithms face when the field is not prepared for automation. Some of these challenges are: uneven distribution of vegetables in the field, intra-species variability, illumination, occlusion and clutter. From a technical point of view, all these aspects translate to covariate shifts and lack of labelled samples. For example, a study on the intra-species covariate shift of sweet peppers and its impact on detection and segmentation algorithm is given in \citep{Halstead2020fruit}. The authors explore the issue of generalisability by considering a fruit that is grown using different cultivars  and in different environments (field vs glasshouse). Their results show how in single task learning the performances drop significantly, as low as 0.323 of F1-score, on cross dataset detection, and only by setting up a multi-task learning problem they are able to increase this score by a good margin, thanks to the multiple back propagation signals. Leveraging the cross-task correlations can be seen in itself as a form of self-supervised learning. 
 Even with these approaches, all detection algorithms need some data of the target distribution to train on, and it is often difficult to collect a good amount of labelled images that catch the actual distribution variability. 

A good example of a crop with a wide range of varietal variability is the grapevine. Wine and table grapes are different in sizes and bunch structure, and the vines are trained in different ways. Even between different varieties of each type of grapes the variability in size, shape, color, foliage and vine structure makes detection related tasks difficult to generalize. 
With respect to these considerations, a number of works are relevant for our discussion. Early approaches to grape detection and counting are characterized by the use of fine-tuned hand crafted features. For example, an early approach of grape detection is presented in \citep{SkrabanekBook2016simplified}. Here the authors use histogram of oriented gradients (HOG) descriptors together with a Support Vector Machine to build a white wine grape detector. An approach that builds on these early results and data is the one presented by \citep{PerezZavalaCEA2018pattern}. The authors use again a solution based on hand crafted features, \ie HOG, fast radial simmetry transform (FRST) and linear binary patterns (LBP), to feed a support vector machine (SVM) based detector, and use geometrical considerations to separate self-occluding grape bunches. The yield estimation task is then a result of the computation of the number of berries detected. Both these solutions show some robustness to color and illumination variability, but require a good deal of tuning of the algorithms, which limits the reuse of trained systems for other varieties.
Another approach to grape yield estimation that is based on geometrical considerations is that of Liu et al. \citep{LiuCEA2017computer}, where the detection is done on the early stage buds that shoot from the branches in an unsupervised fashion only by using Gaussian fitting. The advantage of this method is the independence from labelled data and the reliance on a simple camera as input. In this sense, this work is close to ours, but the approach is usable only in the early stage season and does not take into account the yield loss for malformed and diseased grape bunches. We consider this kind of approach complementary to ours, since it can be used to have early season prediction that can be refined later by proper detection based methods.

More recently, a good number of deep learning based detection algorithms have been proposed. 
\citep{Palacios2022early} present an early yield prediction system based on berry counting using a SegNet segmentation network to extract features from the bunches and canopy images, such as the number of estimated visible berries, or the ratio of leaves in the detection bounding boxes. The statistics on the variability of the features of the 6 grapevine varieties stresses how cross-variety distribution shifts are significant for learning techniques. The authors show a normalized root mean sqared error (RMSE) of $23.83\%$ on their data between the counted and real number of berries. However, the whole system is still run at night with direct illumination.
Another berry detection system based on CNNs is presented in \citep{Zabawa2019detection} and in the extended work \citep{Zabawa2020counting}. In these works, the authors reach a very good $94.0\%$ and $85.6\%$ for the two grapevine training systems considered, but again the images were collected with a modified straddle grapevine harvester for illumination constancy. This solution is then not applicable to, for example, table grapes, for which a straddle machine is non viable. 
In \citep{Coviello2020gbcnet}, a counting network inspired by density based crowd counting techniques was presented. The authors show very interesting results,  with mean average error (MAE) ranging from $0.85\%$ for Pinot Gris variety, to $11.73\%$ for Marzemino variety. To achieve to these results, they had to label more than 35000 berries, a very time consuming operation. 
All these approaches have the common characteristics of leveraging deep learning for increased performances, but limiting the variability of the problem by using specific machines of huge quantities of labelled data. These two aspects limit the applicability of such techniques to the specific case for which they were conceived. 

The aforementioned considerations inspired us to explore methods that could help in training detection and instance segmentation algorithms with few labelled data. We explicitly consider the case where a small amount of labelled data from a similar cultivation has been collected and labelled (Source Dataset, SD), but which is not enough to get acceptable detection and segmentation performances on a different orchard with consistent covariate distribution shift (Target Dataset, TD). We use as our test target data example table grape vineyards cultivated in Aprilia, southern Lazio, while our source dataset is the Embrapa Wine Grape Instance Segmentation Dataset (WGISD) presented in \citep{SantosCEA2020grape}. 

We present a combination of weakly and semi supervised techniques that are able to significantly increase the performance of the algorithms and we compare these newly trained algorithms with the state-of-the-art approaches on the example application of tracking fruits for yield estimation. The proposed solution is able to produce pseudo labelled data in order to bridge the gap of covariate shifts that occur whenever a new specific crop becomes the target of a computer vision system for precision agriculture. We explicitly tackle the problem of doing so with limited hardware and software resources, to address the needs of small and medium businesses. For this reason, all the pseudo labelling strategies presented are based on simple videos collected with a cellphone camera. 

With this in mind, the specific pseudo labelling strategies we propose are of two kinds:
\begin{itemize}
    \item Automatic bounding boxes generation for objects contained in consecutive video frames, based on a starting estimate and 3D structure geometrical considerations. We show that, leveraging a simple initial labelling - which could be manual or automatic - and the information that we can get from feature matching and structure from motion, we are able to generate new labelled data that greatly increases the performance of the detector. 
    \item Pseudo mask generation for instance segmentation: we show how, starting from a simple bounding box -~which could be the one automatically generated in the previous step~- it is possible to use a segmentation network together with a refining strategy to generate new mask labels. 
\end{itemize}
Self-supervised techniques have been frequently proposed to solve the data scarcity problem in specific scenarios (see, for example, \citep{Granland2022detecting}, \citep{Li2022leaf} and \citep{Siddique2022self-supervised} for some recent detection or segmentation approaches), however, to the best of the authors' knowledge, this is the first time that a general self-supervision technique for detection and segmentation in agriculture is proposed in order to tackle a whole category of problems.
    
The structure of the paper is the following. In Section \ref{sec:m&m}, the pseudo-label generation system (PLG) is detailed, both for the detection (Section \ref{sec:upper_subsystem}) and for the segmentation tasks (Section \ref{sec:lower_subsystem}). In addition, a tracking task for yield estimation is discussed in Section \ref{sec:tracking}. In Section \ref{sec:dataset}, we describe the data collected and used, and in Section \ref{sec:metrics} the specific metrics for multiple object tracking (MOT) are defined. Experiments and discussion for all these systems are described in Section \ref{sec:experiments} and conclusions are drawn in Section \ref{sec:conclusions}.

\definecolor{myred}{rgb}{0.68, 0.33, 0.33} 
\definecolor{mylightred}{HTML}{DE8787} 
\definecolor{myorange}{rgb}{1.00, 0.60, 0.33} 
\definecolor{mygreen}{rgb}{0.17, 0.63, 0.17} 
\definecolor{mycyan}{rgb}{0.00, 1.00, 0.80} 
\definecolor{myblue}{HTML}{00CCFF} 

\section{Materials and Methods} \label{sec:m&m}
In this Section, we discuss the data, the general architecture of the system, and the algorithms on which it is based. We start by describing the experimental field where the target data has been collected, and then  how it was collected, and how it compares to the source data that was already available. Then, we describe the global system architecture and introduce its components. The final sub-sections introduce the metrics used for our experimental evaluation, and give more details of each subsystem. 

\subsection{Experimental Field} \label{sec:field}
\begin{figure*}[t]
\centering
	\begin{subfigure}[]{0.24\textwidth}
		\includegraphics[width=\columnwidth]{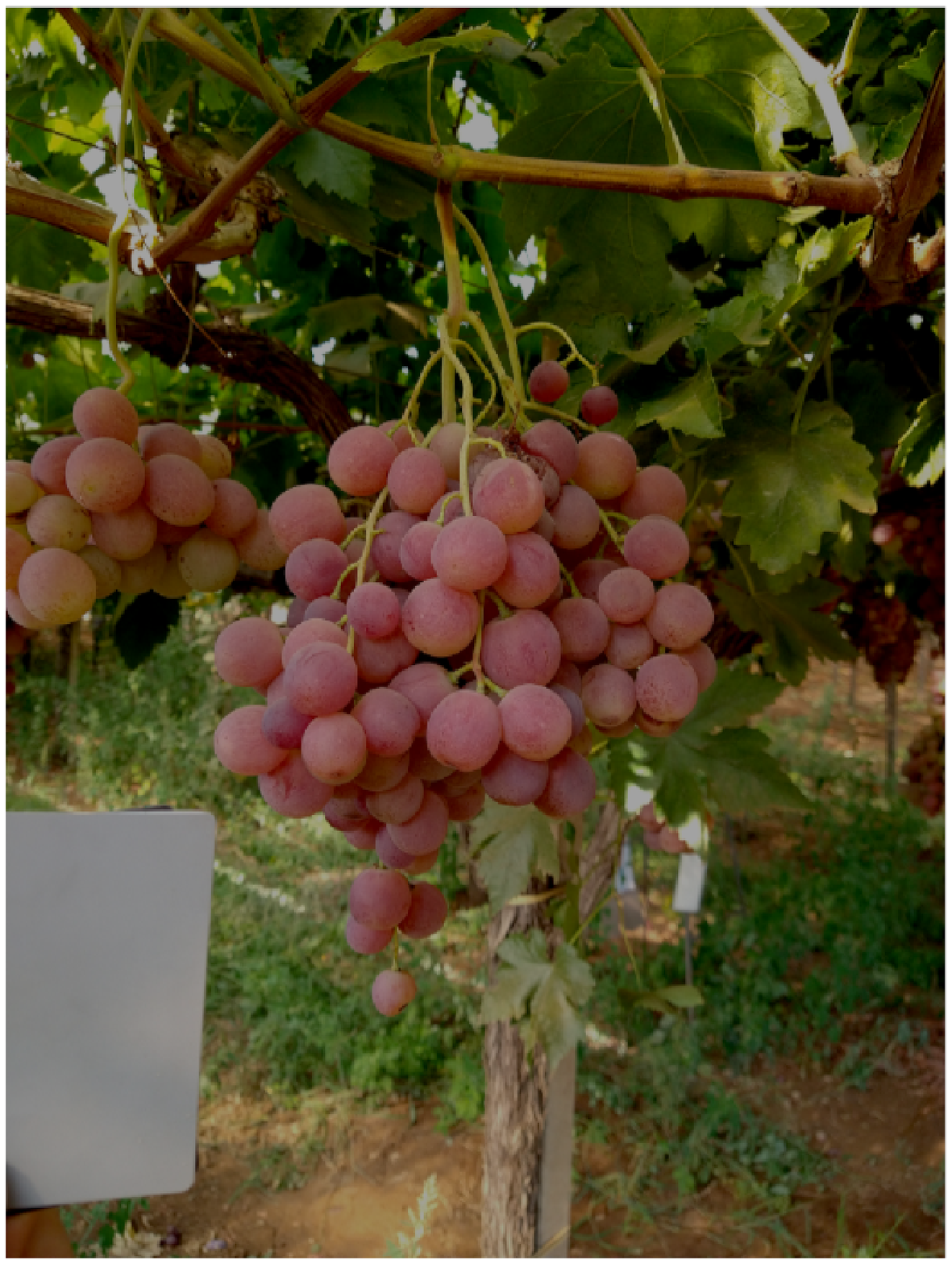}
		\caption{}
	\end{subfigure}
	\begin{subfigure}[]{0.24\textwidth}
		\includegraphics[width=\columnwidth]{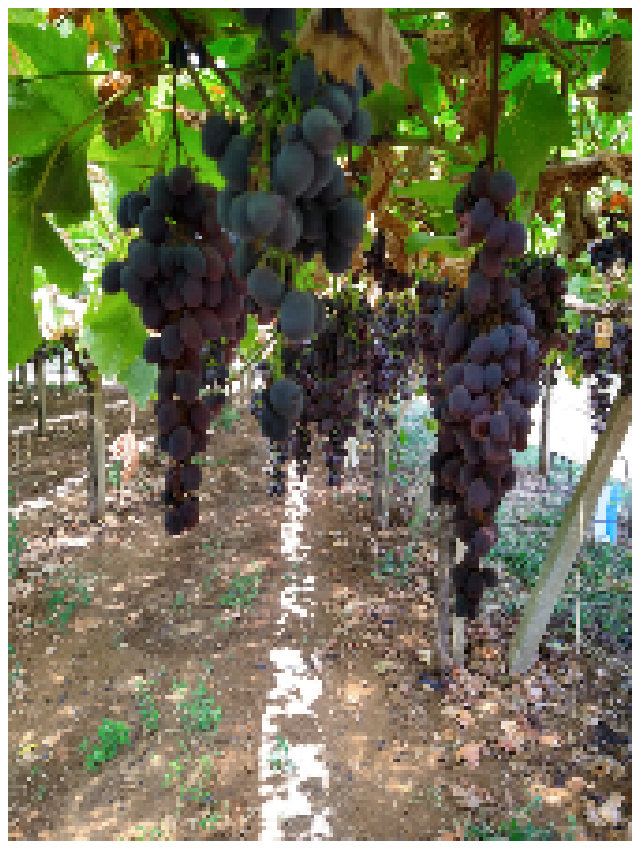}
		\caption{}
	\end{subfigure}
	\begin{subfigure}[]{0.24\textwidth}
		\includegraphics[width=\columnwidth]{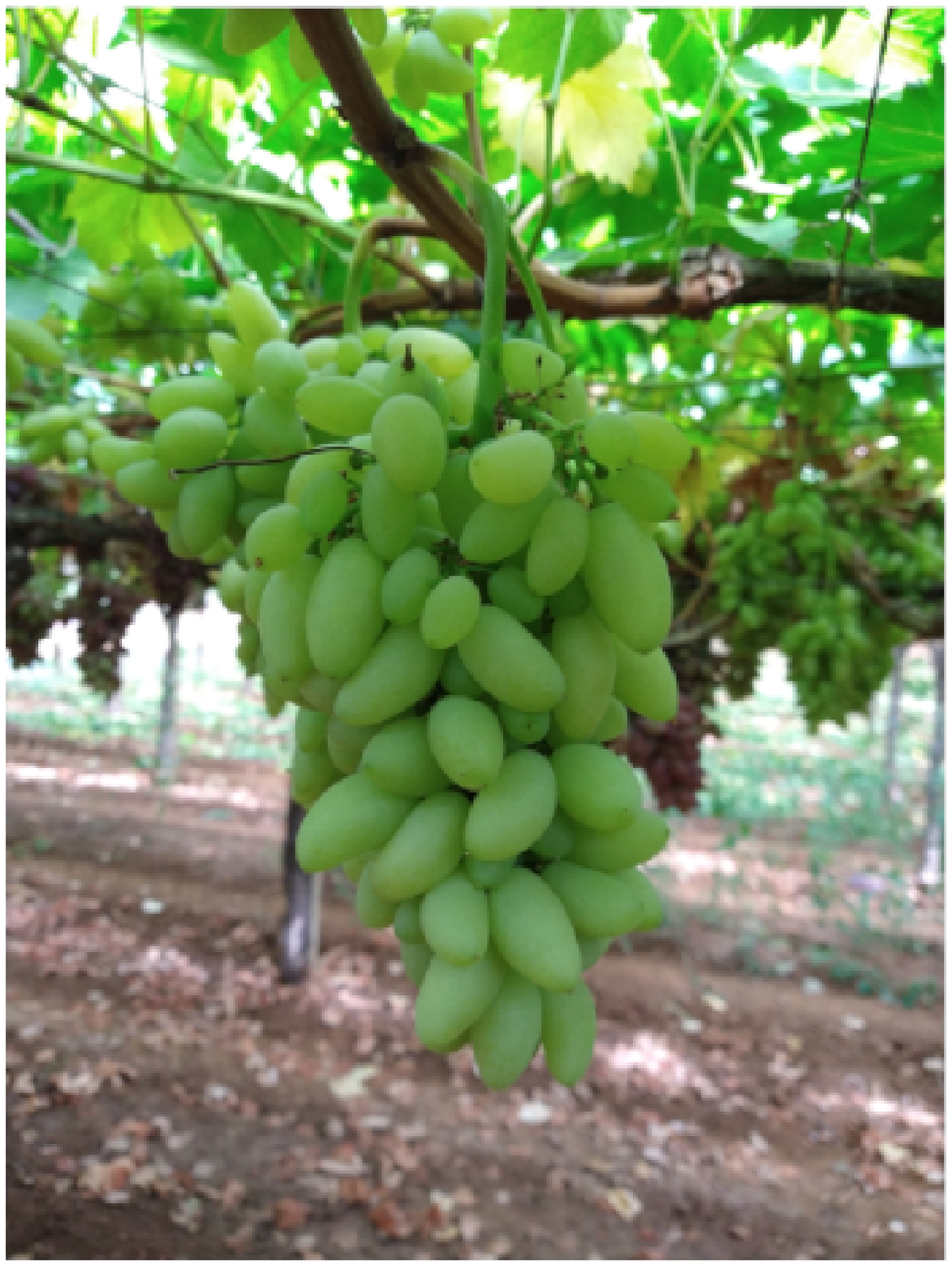}
		\caption{}
	\end{subfigure}
	\begin{subfigure}[]{0.24\textwidth} \label{fig:black_pizzutello}
		\includegraphics[width=\columnwidth]{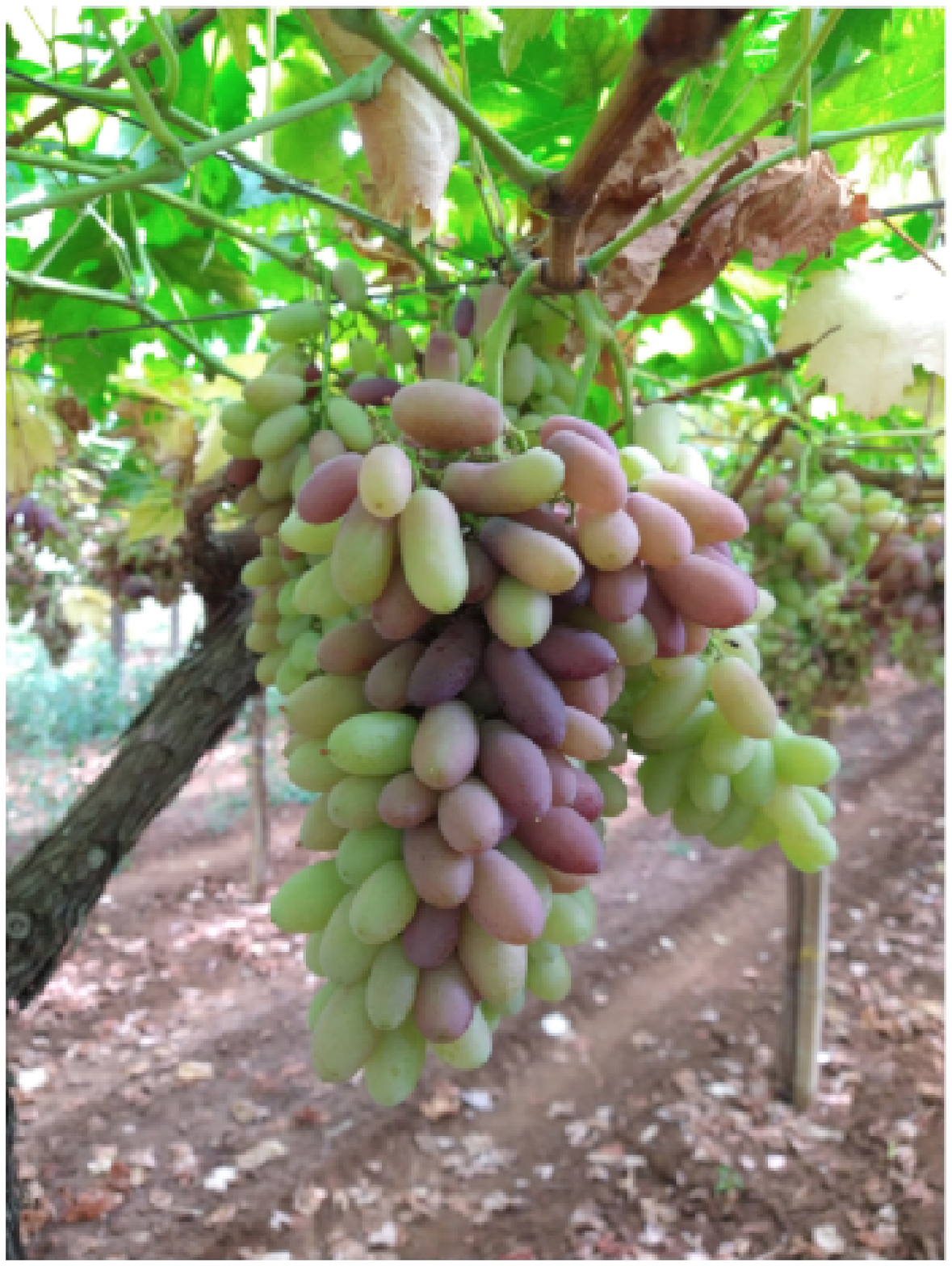}
		\caption{}
	\end{subfigure}
	\caption{An example of the four varieties present in the experimental field. The Black Pizzutello (d) is the most interesting for this work because it presents the highest variability in shape and color with respect to other rounded berry variants.}
\label{fig:grapes}
\end{figure*}

\begin{figure*}
\centering
	\begin{subfigure}[]{0.3\textwidth}
		\includegraphics[width=\columnwidth]{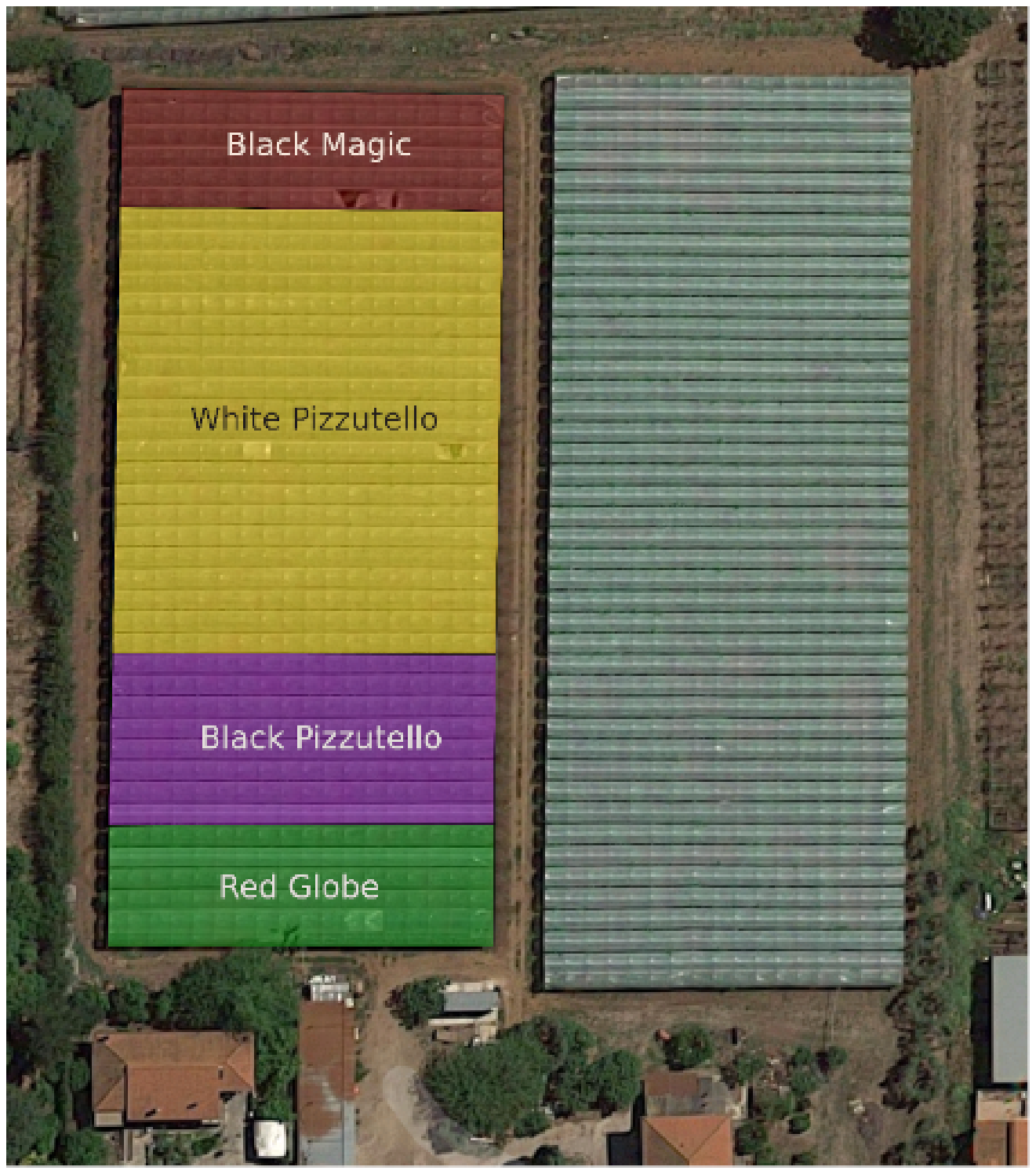}
		\caption{}
	\end{subfigure}
	\begin{subfigure}[]{0.605\textwidth}
		\includegraphics[width=\columnwidth]{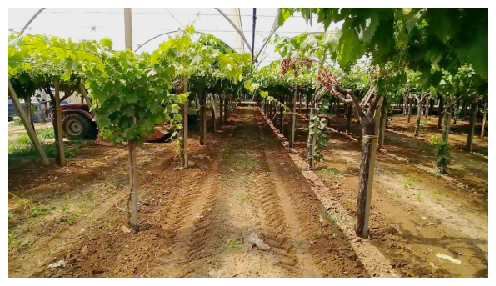}
		\caption{}
	\end{subfigure}
	\caption{a) Satellite view of the experimental vineyard. The image shows the varieties that are grown on each row. b) A picture of the trellis (Tendone) structure.}
\label{fig:field}
\end{figure*}

The experimental field is located in southern Lazio (Italy). The vineyard is composed of two plots approximately 114 m x 51 m (0,58 ha) and 122 m x 48 m (0,58 ha). 
Vineyards are structured as a traditional trellis system called \textit{Tendone} with a wide distance between each plant, 3 x 3 $m^2$. Plantations are all older than 3 years and so in full production and health, thus representing a typical working condition for the validation of agronomic activities such as fruit harvesting or vine pruning. 
All structures are traditionally covered with plastic and net to protect grapes from rain and hail. The average extension of each plot is around 1 hectare and dimensions (length and width) are on average between 25 m and 50 m according to plot extension and geometry. The selected vineyard in Aprilia has currently four different table grape varieties which are described in the following: White Pizzutello, Black Pizzutello, Red Globe and Black Magic. Figure \ref{fig:grapes} shows some examples of these grape varieties while Figure \ref{fig:field} shows images of the experimental field as well as the approximate extension of each grape variety in the vineyard. 

Of the four varieties that were present in the vineyard, Black Magic was of very low quality and thus untended by the field owner. White Pizzutello is identical in shape to the Black one, and the latter has the same color as the former when not ripe. Together, white and black Pizzutello are a peculiar variety of the Lazio Region and present the highest variability in shape and color with respect to standard rounded berry variants. For these reasons, while we collected images of all the varieties, we finally concentrated our data labelling effort only on Black Pizzutello.  

\subsection{Dataset} \label{sec:dataset}
\begin{figure*}[h]
\centering
	\begin{subfigure}[]{0.48\textwidth}
		\includegraphics[width=\columnwidth]{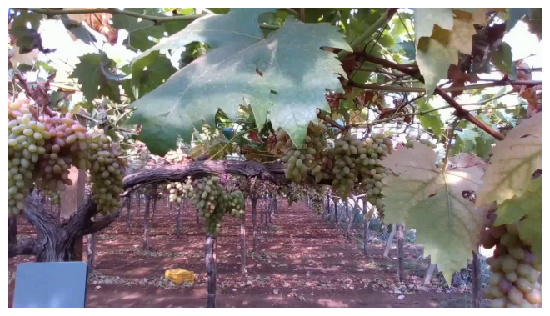}
		\caption{}
	\end{subfigure}
	\begin{subfigure}[]{0.48\textwidth}
		\includegraphics[width=\columnwidth]{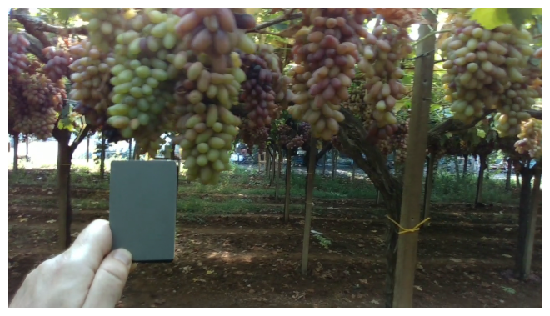}
		\caption{}
	\end{subfigure}
	\begin{subfigure}[]{0.24\textwidth}
		\includegraphics[width=\columnwidth]{images/pizzutello_nero.eps}
		\caption{}
	\end{subfigure}
	\begin{subfigure}[]{0.48\textwidth}
		\includegraphics[width=\columnwidth]{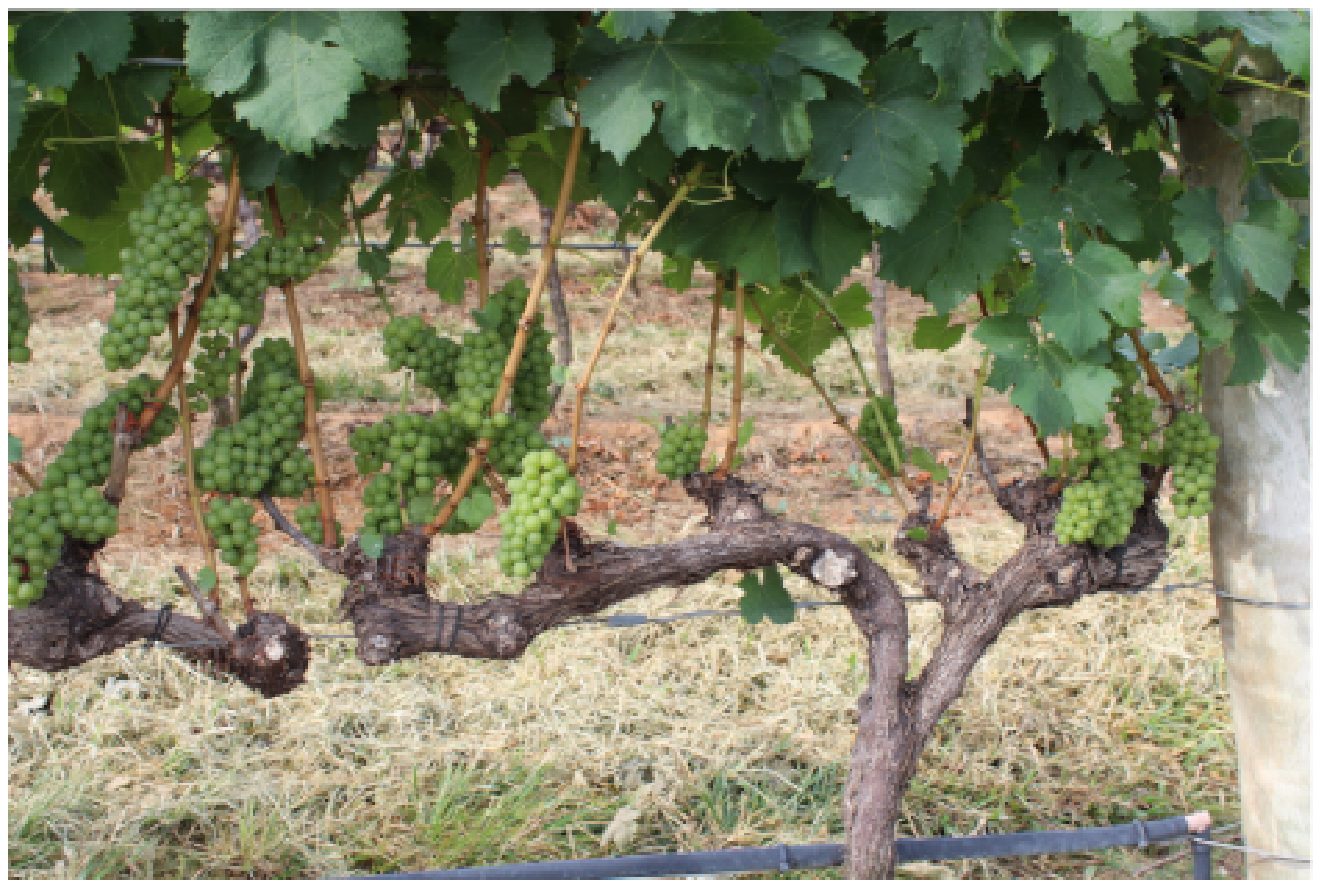}
		\caption{}
	\end{subfigure}
	\caption{a) and b) Examples of the video frames collected for our target dataset (TVid): we stress that these are simple cellphone camera based videos and are the only new data required for the system to be able to produce new labels without supervision. c) Example of TImg dataset. d) Example from the source dataset, the WGISD dataset \citep{SantosCEA2020grape}. It is possible to note the differences in shape, color and general illumination conditions.}
\label{fig:video_examples}
\end{figure*}

As we mentioned in the introduction, the proposed system deals with the covariate shift from a generic source dataset to a target dataset that is representative of the images that could be collected on the field. 
We assembled our target dataset with two different kinds of data. The first are videos recorded using a mid range cellphone camera (MotoG8 Plus), which simulates a data collection operation that could be performed by a farmer with ease. We collected videos moving along the vineyard (\ie tangential to the rows), without any requirement on distance from the fruits or height from the ground. In this work, we use HD (128$\times$720) videos at 10Hz with a total of 1469 frames. Examples of these frames are shown in Figure \ref{fig:video_examples}. A short segment of 10 seconds has been labelled for test use in the case of the tracking algorithm evaluation, while the rest has been used without labelling thanks to the semi-supervised nature of the system. We briefly call this target video dataset TVid. Note that we collected the images using a neutral grey card for white balance purposes, but this is meant for future uses, and it is not a requirement for the methods presented in this work.

The second kind of data is composed of static images of Black Pizzutello. This data simulates the images a farmer, or a robot, could collect to perform some agricultural action on specific grape bunches (\eg quality estimation, disease detection, automatic harvesting). The dataset consists of 134 images of 3000x4000 resolution, collected with the same cellphone camera used for the videos, however the optics and chip used for video and still images are different, as is often the case with cellphones. This is intentional, since it adds a very common source of covariate shift related to the device and capturing mode (motion vs still images). All the images in this case have been labelled for detection (bounding boxes), while a small subset has also been labelled for instance segmentation (70 images), using the Innotescus labelling application \citep{Innotescus}. All these labels are used for validation and testing of the algorithms described in this Section. We call this still images dataset TImg. Together these datasets (TVid and TImg) constitute our Target dataset (TD).

As mentioned above, we work under the hypothesis that a small amount of labelled data of the same fruit exists, but that it has considerable covariate shift with respect to the TD distribution. In this work, our Source Data is the one presented by Santos \etal \citep{SantosCEA2020grape}. For the details about these data, the reader can check the cited work. Here we give a short summary to underline the differences between SD and TD in terms of:
\begin{itemize}
\item the grape varieties (wine vs table, berry shape and color)
\item the illumination conditions (full sun vs shadows)
\item the camera device (Reflex vs cellphone camera)
\item scale of the images (standard scale vs variable scale)
\end{itemize}
 To quantify the covariate shift gap in Section \ref{sec:experiments}, the performance drop for detectors and instance segmentation networks trained on SD and tested on TD are given. 

\begin{figure*}[h]
\centering
\includegraphics[width=\textwidth]{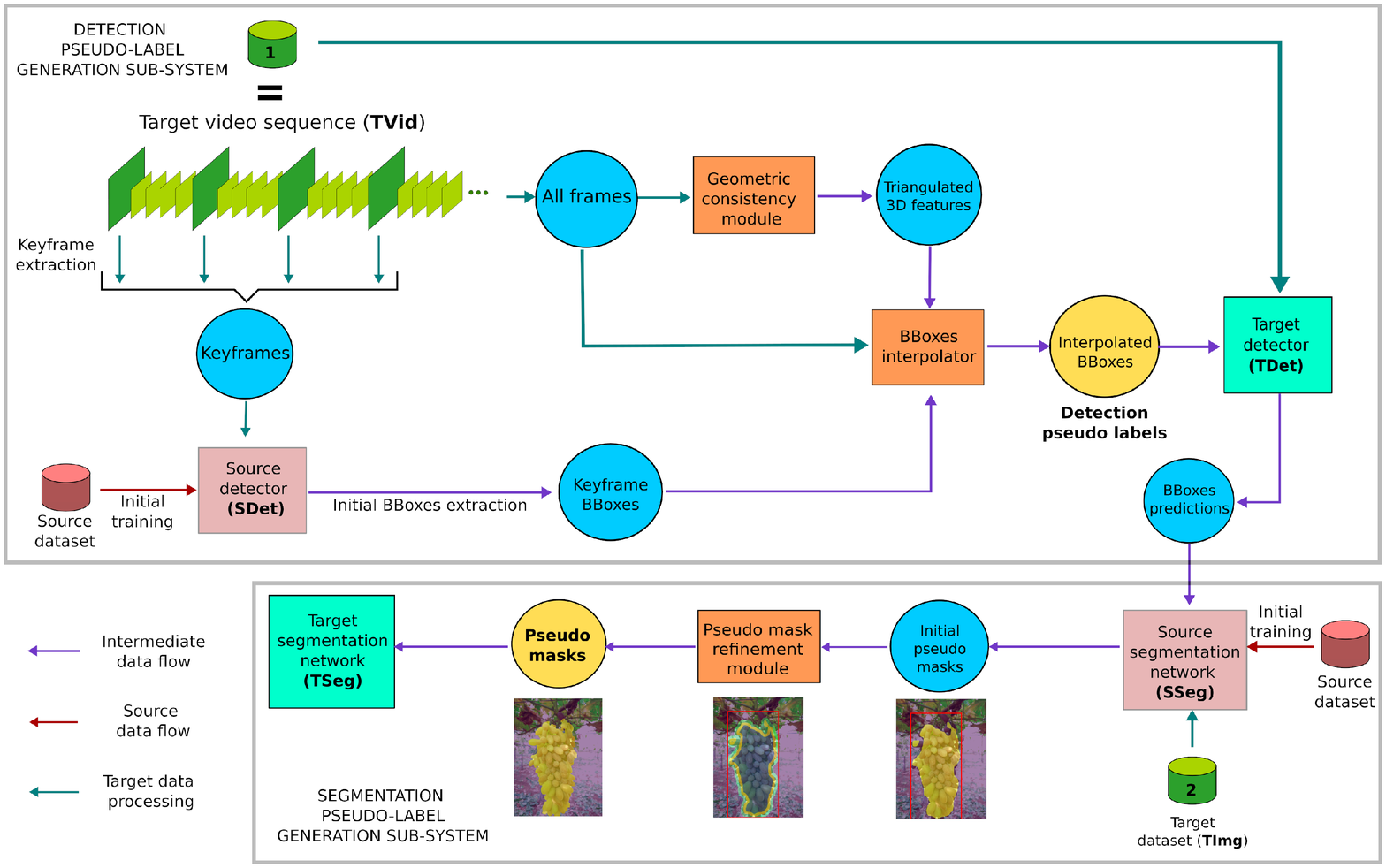}
\caption{This figure shows the complete system architecture. The inputs are a source dataset (\textcolor{myred}{\textbf{red}} cylinder) and a video collected on the field by the robot or a farmer (first \textcolor{mygreen}{\textbf{green}} cylinder, TVid, expanded as a sequence of frames to show the keyframe selection process). SDet and SSeg (\textcolor{mylightred}{\textbf{light red}} blocks) are the initial detection and segmentation networks trained only on the source dataset. All the intermediate computing blocks are depicted in \textcolor{myorange}{\textbf{orange}}, while the intermediate outputs are in \textcolor{myblue}{\textbf{blue}} circles. Both the pseudo bounding boxes and pseudo masks produced are depicted in yellow, while the detection and segmentation networks trained on these new labels (TDet and TSeg) are depicted in \textcolor{mycyan}{\textbf{light green}}. The data flow in the system is also color coded as per legend. }
\label{fig:sys}
\end{figure*}

\subsection{System Overview} \label{sec:overview}
An overview of the system is depicted in Figure \ref{fig:sys}. 
The main inspiring principle of this work is the economy of data labelling and data reuse. For this reason, the only two sources of data are the source dataset (data available from a similar task) and a video collected on the target field. The source dataset is used to train the initial detection and segmentation models, namely the Source Detector Network (SDet) and the Source Segmentation Network (SSeg). SDet is not perfectly tuned on the target environment, still it can be used on selected frames of the video input that we call keyframes. To keep this solution simple, we consider equally spaced keyframes starting from the first one, but other strategies could be devised. A set of initial bounding boxes is extracted from this keyframe, using a high confidence threshold, to limit the false positives. Then, the whole video is passed in a Geometric Consistency block (GC block) that extracts features from each frame and associates them. We tested two different options for this block, as will be shown in the following sections. Using this geometric information, together with the initial bounding boxes extracted from the keyframes, it is possible to interpolate the bounding boxes positions for the remaining frames with high accuracy. These new bounding boxes are our pseudo-labels for training the detector on the target environment, which we call Target Detector (TDet). 

The Detection Pseudo-Labels Generation (DPLG) sub-system could be used independently by the Segmentation Pseudo-Labels Generation (SPLG). To prove the effectiveness of the approach, we compare the performance of TDet on the bunches tracking problem, \ie counting the number of grape bunches by counting the instances tracked along a video. This problem is relevant since it can be used for yield estimation purposes. We test two different tracking algorithms, which are described in Section \ref{sec:tracking} and evaluated in Section \ref{sec:tracking_results}.

The goal of the second part of the system is to generate pseudo masks for training an instance segmentation network. This sub-system can be seen both as an independent pseudo label generator, or as part of a bigger system such as the one we describe here. As mentioned before, the SSeg is trained only on source data and is not able to produce good segmentation masks on the Target Data. However, it is possible to give the network some information cues that can greatly improve the mask estimates. The first one is the bounding box region in which the instance should be segmented. This cue comes easily from the previous step of pseudo bounding box generation, but it could be produced otherwise. This generates the initial pseudo masks. It is possible to use these pseudo masks for training the Target Segmentation Network (TSeg) but this would lead to poor performances due to \textit{confirmation bias}. We need therefore to inject external information from other cues that we have. In our system, this is the role of the pseudo masks refining block. In Section \ref{sec:refining}, three different solutions for refinement will be described. Thanks to these refined pseudo masks it is finally possible to train the TSeg Network. Section \ref{sec:segmentation_results} reports the results of the refining strategies and compares the performance of TSeg with SSeg.

Finally, the experiments of the whole system, trained only on videos and tested for instance segmentation performance, are given in Section \ref{sec:joint_sys_results}.

\subsection{Metrics} \label{sec:metrics}
In this Section, we describe the metrics used to evaluate and compare the detectors, the trackers and the instance segmentation algorithms. 
To evaluate the detectors and instance segmentation algorithms, the standard metrics of Precision, Recall and Intersection over Union (IoU) have been used. In addition, for instance segmentation, the Average Precision, as defined in the MS COCO challenges \citep{Lin2014microsoft}, has been used. 

Usually AP is computed for each class and then averaged to obtain the mean average precision (mAP). In this work, since there is only one class (grape), the AP coincides with the mAP.
In the MS COCO metrics, the AP is calculated by computing the precision at every recall level from 0 to 1 with a step size of 0.01. The mAP is then computed by averaging the AP over all the object categories and ten IoU thresholds from 0.5 to 0.95 with a step size of 0.05. 

To evaluate the trackers, we follow the common practice of Multiple Object Tracking (MOT) as defined by Wu and Nevatia \citep{classicmot} and the CLEAR MOT metrics \citep{clearmot}. MOT is a difficult task to evaluate, since the performance metrics should capture both the precision in detecting individual instances and the accuracy in tracking each instance across multiple frames, without losing track or switching between instances. Given a number of objects $o_j, j \in [0..m]$, the tracker produces a number of hypotheses $h_i, i \in [0..n]$. The performance of association of hypothesis and objects can be measured frame by frame by using the classic True Positive, True Negative, False Positive and False Negative figures, together with their direct descendants Precision and Recall. However, recently a number of compound indexes have been proposed to better capture the general tracker's performance. The first one is the Multiple Object Tracking Accuracy (MOTA), defined as follows:

\begin{equation}
    \label{eq:mota}
        MOTA = 1 - \frac{(FN + FP + ID_{sw})}{GT} \in (-\infty, 1]
\end{equation}
where $FN$ and $FP$ are False Negatives and False Positives, $ID_{sw}$ represents the number of instances whose ID has been erroneously switched, GT is the real number of instances in the video. This index accounts for three sources of error, namely the false positive ratio, the false negative ratio and the mismatch ratio. Together, they give an idea of the general tracking accuracy. 
To evaluate the precision, a second index was proposed:
\begin{equation}
    \label{eq:motp}
        MOTP = \frac{\sum_{t,i}{d_{t,i}}}{\sum_{t}{c_t}}
    \end{equation}
where $c_t$ denotes the total number of matches in frame $t$, and $d_{t, i}$ in general represent the distance of the hypothesis and the object, but in our case can be computed as the overlap of the ground truth and the hypothesis bounding boxes. This second index gives only a measure of the precision in detecting the instances without giving any information on the tracking and association capability.  
    
\subsection{Detection and Segmentation Network Architectures} \label{sec:sdet_sseg}
As explained in Section \ref{sec:overview}, the general pseudo label generation system is based on pre-trained detection and segmentation networks (SDet and SSeg) and is meant to produce the pseudo labelled data to train new networks that are able to perform better on TD (TDet and TSeg). 

The main parameters that influence the choice of the architectures are speed and accuracy. It is well known (\citep{Liu2020deep}) that SotA detection networks can be divided into two main categories: two stage and single stage. The first kind separates detection into two phases, the first is called \textit{region proposal} and gives object bounding boxes candidates, while the second filters and refines these candidates to produce the bounding boxes and classifies the objects. The second kind instead extracts both region proposal and class prediction in one pass. The main advantage of the single stage detectors is speed, which is much higher than the two stage one, but at the cost of a general reduced accuracy. The main examples of single stage detectors are the YOLO variants, in particular the recent YOLOv5 \citep{redmon2018yolov3}. One of the best known and best performing two stage architectures is Mask R-CNN \citep{He2017ICCVmask}, which is also a segmentation network, more accurate than any YOLO variants, but slower and difficult to tweak for real-time use.

In this work, we use the single stage YOLOv5s architecture for the experiments on tracking, since real-time detection is needed for this kind of application. In addition, some of the variants have a small number of parameters, which makes them viable for embedded applications, such as robotic harvesting.
The pseudo bounding box generation could be performed offline, thus allowing use of the better performing Mask R-CNN, but we decided to use the YOLO detector to keep this sub system self-contained. In addition, using an architecture with lower detection performance stresses and tests the robustness of the generation process. 
For segmentation and pseudo mask generation, a segmentation network is needed, so the choice falls on Mask R-CNN. The details of the pretraining and fine tuning of the detection and segmentation networks are given in Sections \ref{sec:sdet} and Section \ref{sec:sseg}, respectively. 

\begin{figure*}[h]
\centering
\includegraphics[width=\textwidth]{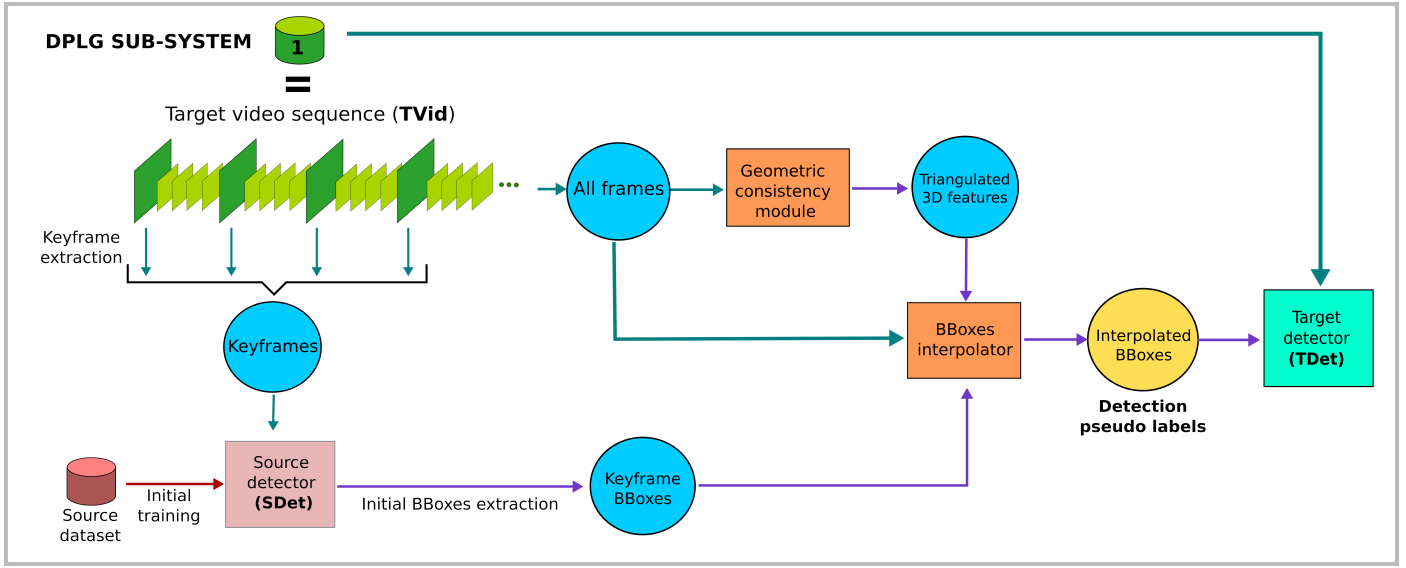}
\caption{This figure shows the detection pseudo-label generation (DPLG) sub-system alone. The source dataset is used to train an initial coarse bounding box detector (SDet) that is then used, together with the SfM system, to generate a large number of new labelled images from the frames of continuous videos of the vineyard. This same system can be applied to other fruits with relative simplicity. }
\label{fig:upper_subsys}
\end{figure*}

\subsection{Detection Pseudo-Label Generation Sub-System and Tracking application} \label{sec:upper_subsystem}
In this Section, we detail the elements of the DPLG system depicted in Figure \ref{fig:upper_subsys}, \ie the pseudo bounding box generation system, together with the tracking algorithm used for yield estimation as a possible application. 

\subsubsection{Geometric Consistency Block} \label{sec:sfm_block}

\begin{figure*}
\centering
	\begin{subfigure}[h]{0.45\textwidth}
		\includegraphics[width=\columnwidth]{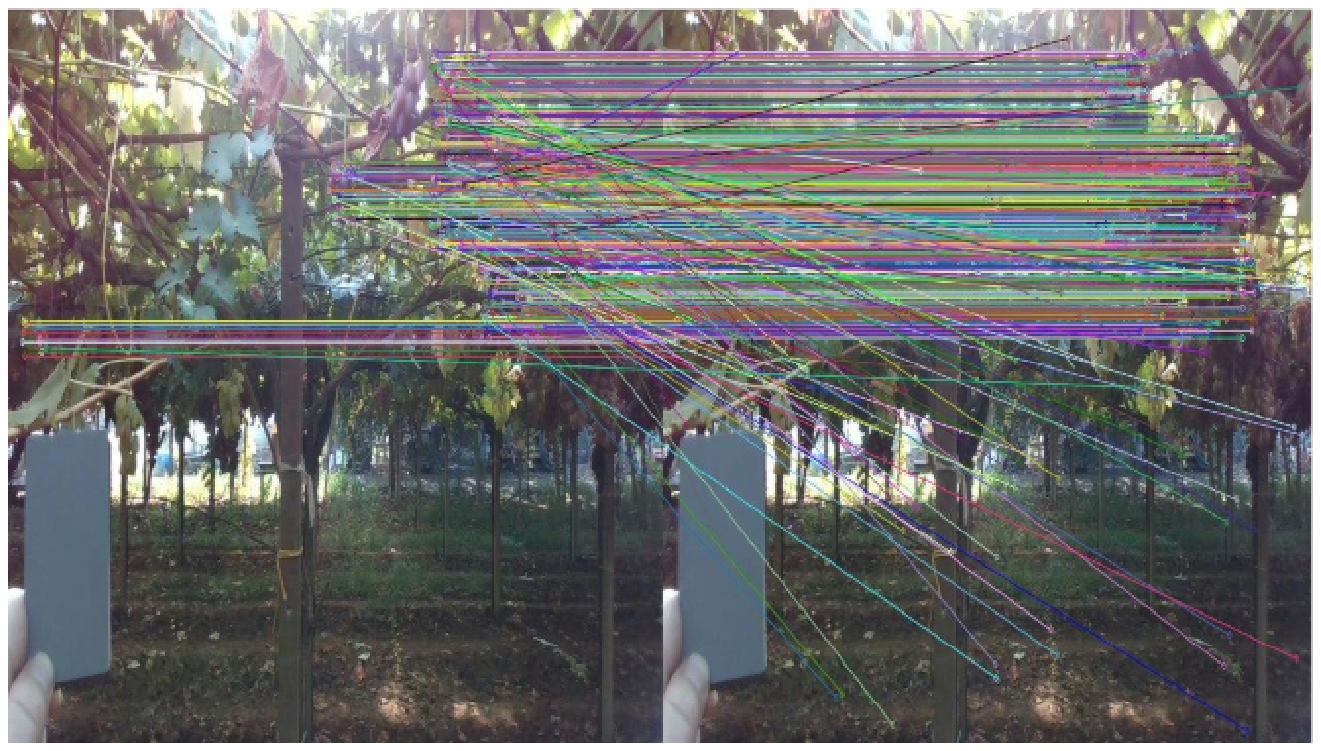}
		\caption{}
	\end{subfigure}
	\begin{subfigure}[h]{0.45\textwidth}
		\includegraphics[width=\columnwidth]{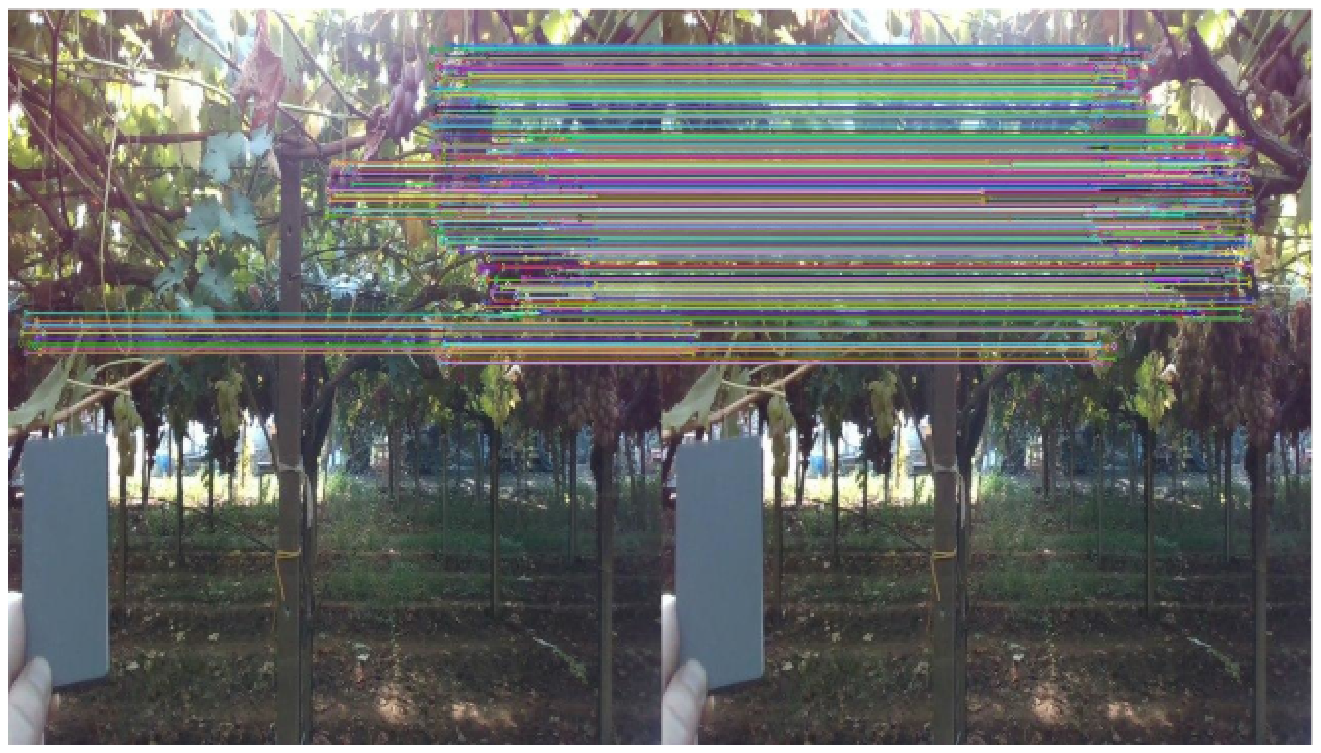}
		\caption{}
		\label{fig:ransac_result}
	\end{subfigure}
	\caption{Feature matching and geometric verification using RANSAC: COLMAP SfM in its lightest form still requires considerable computational time and it is viable only for offline elaborations, while 2D feature matching requires much less computation and potentially runs in real time. a) The matching of surf features with brute force matching b) the same matching refined with homography computation combined with RANSAC selection.}
\label{fig:matching_example}
\end{figure*}

The purpose of this block is to use geometrical correspondences extracted through epipolar geometry to associate grape instances in different frames of a video stream, \ie, given a detected grape bunch, by identifying 2D features belonging to it and matching, or triangulating, them across multiple frames, it is possible to find the same bunch instance in the following frames. We use this strategy in two ways in this work, first to extract pseudo bounding boxes, and then for tracking. In this Section, we describe the general functional principles of SfM algorithms and their computational costs. 

We experimented with two approaches, the first is the same used in \citep{SantosCEA2020grape}, which leverages a SfM software application, namely COLMAP \citep{schoenberger2016mvs, schoenberger2016sfm}. Since SfM is a well known problem, the interested reader can find details of the solutions in \citep{Hartley2006BOOKmultiple, Szeliski2022BOOKcomputer}. In brief, we used the COLMAP modality that extracts sparse features from each frame and then runs a sequential all versus all search and matching of the features extracted from the video. These correspondences are then used to triangulate the 3D points by minimizing the 3D to 2D reprojection error. However, the nature of the problem is such that even with the sparse setting, the computational costs increase exponentially with the number of frames. Our experiments required 5 hours of computation for videos of 500 to 600 full HD frames, on a computer equipped with an Intel-Core i7 3.4 GHz, a Nvidia GTX 950m and 16 GB of memory.

The second approach we experimented addresses this aspect in order to have a real time solution that can be run on an online tracker, such as the one that will be described in Section \ref{sec:tracking}. The idea is that in our context a full SfM solution (\ie, the 3D position of the extracted 2D features in a world reference frame) is not needed, since the kind of videos that are collected in the vineyard are simple walks without closed loops. This means that each table grape bunch is present, at most, in a few consecutive frames, except for the occasional occlusion. For this reason, we found that extracting 2D features from a frame $i$ and from a small number of subsequent frames $i+1, \dots, i+n$, and then matching them was enough to map the grape instance correspondences along the video stream. The features and descriptors used are SURF \citep{Bay2006ECCVsurf}, while the matching is a simple brute force (all-vs-all) distance computation between the corresponding feature descriptors. Since these initial match proposals contain some mismatched features, a RANSAC verification step is used to filter them out. Given that for small camera motions, image transformation could be approximated by a homography transformation, we repeatedly random sample four matches, compute the relative homography, and check how many other matches are correctly predicted by it. The homography with the highest consensus is selected and all matches whose displacement is not compatible with the selected homography, are discarded. An example of this process is given in Figure \ref{fig:matching_example}. The parallel lines left in Figure \ref{fig:ransac_result} show the matches that agree with the homography estimated through RANSAC. Using this approach, even using brute force matching, we were able to reach, without particular optimizations and working only on CPU, 3 frames per second on the aforementioned video and hardware. 

\subsubsection{Bounding Box Interpolation and Pseudo Label Generation} \label{sec:bbox_interpolation}
\begin{figure*}[!h]
\centering
\begin{subfigure}[t]{0.45\textwidth}
    \includegraphics[width=\textwidth]{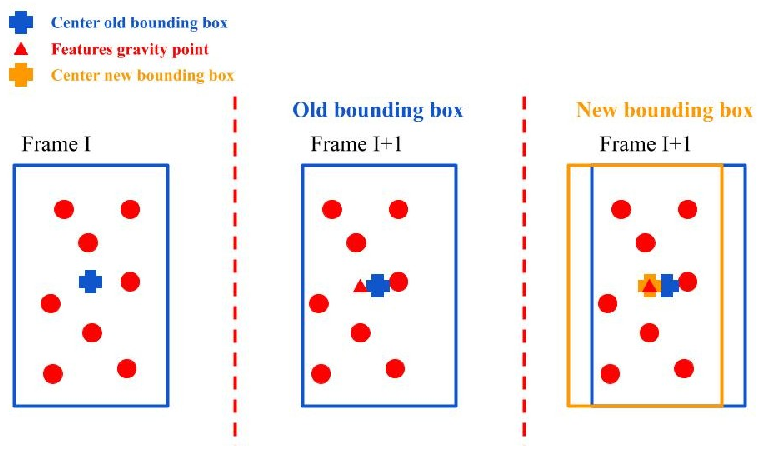}
    \caption{}
    \label{fig:bbox_update}
\end{subfigure}
\begin{subfigure}[t]{0.45\textwidth}
    \includegraphics[width=\textwidth]{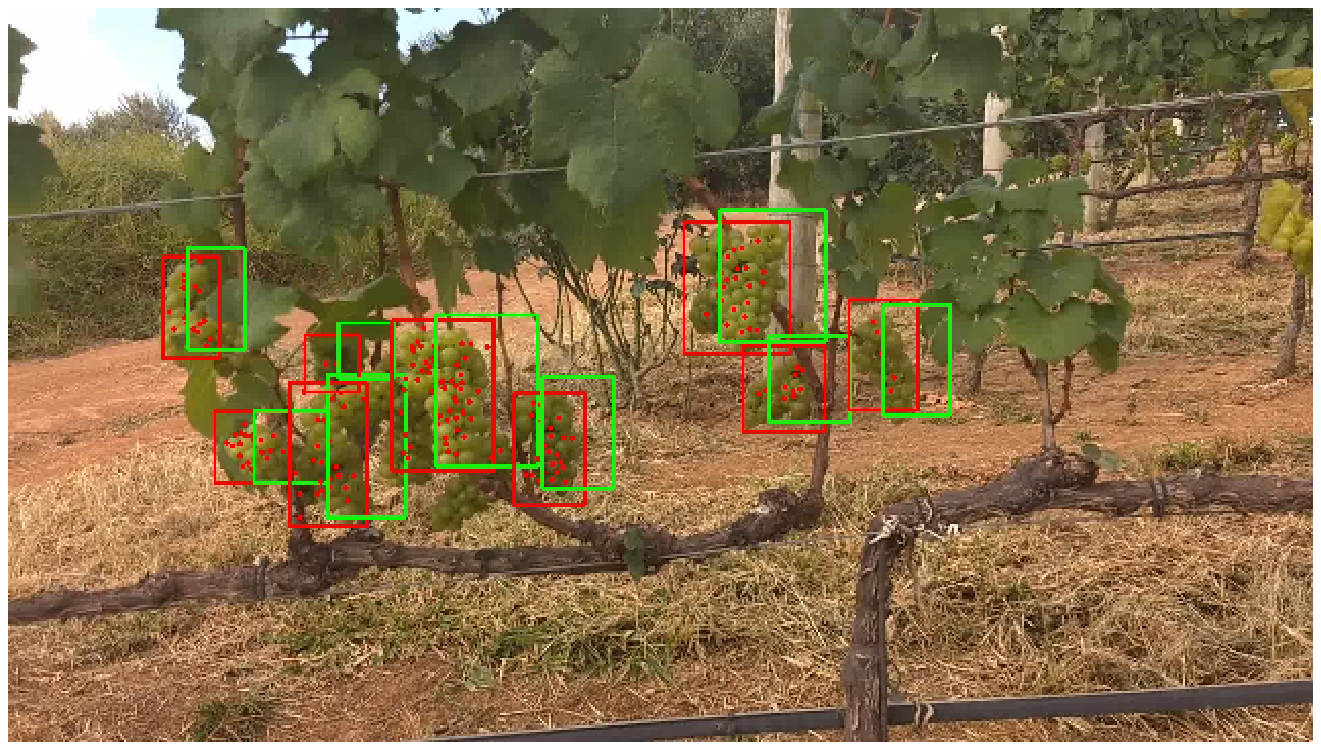}
    \caption{}
    \label{fig:pseudosfm}
\end{subfigure}
\caption{The bounding box interpolation process. a) Shows the updating principle: the bounding box (blue) predicted in frame $i$ is moved to frame $i+n$; while the size remains the same, the position of the new bounding box (orange) is updated by computing the new center of gravity of the features extracted and making it coincide with the box center. b) Pseudo-labels generated by means of the SfM algorithm: the green boxes are the predictions produced by SDet at frame $i$ transposed in the current one ($i+n$), while the red boxes are interpolated ones according to the features matched (represented as red points).}
\label{fig:bbox_interpolation}
\end{figure*}

Bounding box interpolation can be better understood by looking at Figure \ref{fig:bbox_interpolation}. 

Starting from a bounding box found by SDet at frame $i$, thanks to the GC Block, it is possible to have an association between the 2D features contained inside the box with some features in frame $i+n$. Since the camera is moving, both the position of the grapes and the illumination conditions in frame $i+n$ will be different, consequently the features matched will have a different position. The question is then how to draw the new bounding box in frame $i+n$. We use the hypothesis, that the camera is slowly moving, and that the motion is tangential to the direction of the vineyard. Thanks to this hypothesis we can assume that the new bounding box will have the same size as the one found in frame $i$. 

The position of the new bounding box is computed by setting the center of the box to coincide with the centre of gravity of the features in frame $i+n$, as depicted in Figure \ref{fig:bbox_update}. 
Another aspect to consider in evaluating the pseudo bounding box generation scheme is the effect of camera velocity combined with frame rate. If the frame rate of the video is high, or the camera velocity is low, the change in view will be minimal, and consequently the information added by such a sample will be minor. For this reason, we considered it useful to explore the effect of the ratio between keyframes and other frames. We call this parameter \textit{skip} value, since it is the number of frames in which the bounding boxes predicted in frame $i$ are interpolated, before taking a new prediction by SDet. Our ablation experiments showed that using skip 1 (\ie using only SDet to produce pseudo-labels) gave lower performance than using skip 2. However, increasing the skip value seems not to give more advantages. this aspect is explored in Section \ref{sec:tracking_results}, where we show on the tracker application the results of using different skip values. 

\subsubsection{Tracking for Yield Estimation} \label{sec:tracking}
Multi Object Tracking of the grape bunch instances is a preliminary step in yield estimation, as it is possible to estimate the number of bunches by counting the number of trajectories tracked by the algorithm.
The main approach to tracking by detection that we consider is the one presented by Santos in \citep{SantosCEA2020grape} which is based on detection and SfM. However, no metrics were given there to formally describe the performances of the approach. Therefore, we replicated the experiments and computed the metrics using as a target the test sequence of TVid, described in Section \ref{sec:dataset} and depicted in Figure \ref{fig:video_examples}. 
In addition, we tested another detection based State-of-the-Art tracker, DeepSORT \citep{Wojke2017ICIPsimple}, designed to work in real-time using a deep association metric. We chose this tracker since the computation involved in estimating even sparse correspondences between the frames using COLMAP \citep{schoenberger2016sfm} requires considerable time and are not feasible for edge or robotic devices. In addition, the computation of the full SfM solution takes a long time and limits the length of the video to a few hundred frames, while for the second approach there is no such limit. 
In Section \ref{sec:experiments}, we compare the tracking solutions using the MOT metrics with the tracking graphs to gain more insights on what the tracker does and how to improve it further. 
An example of these graphs is given in Figure \ref{fig:id_switch}.

\begin{figure*}[h]
\centering
\includegraphics[width=\textwidth]{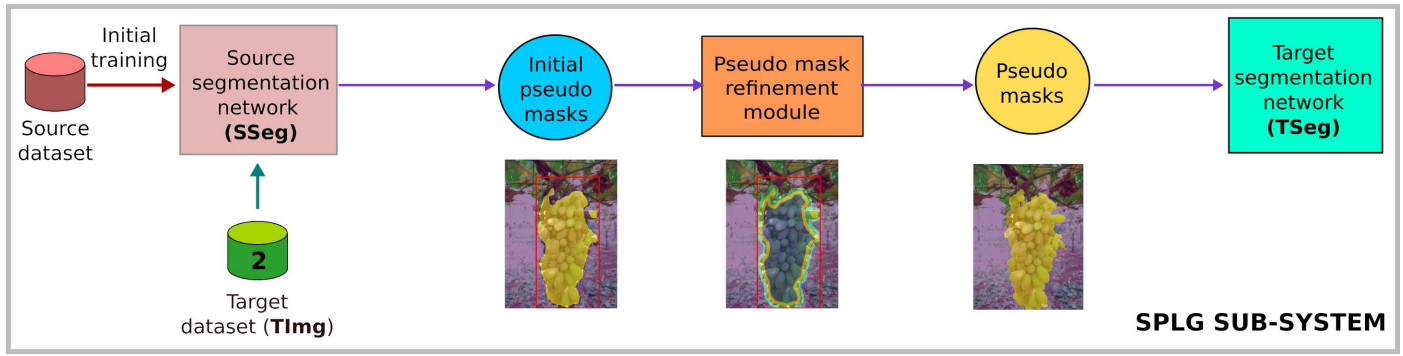}
\caption{This figure shows the segmentation pseudo-label generation (SPLG) sub-system. }
\label{fig:lower_subsys}
\end{figure*}

\subsection{Segmentation Pseudo-Label Generation Sub-System} \label{sec:lower_subsystem}
While detection is enough for counting tasks, for quantitative yield estimations or for tasks that require a physical interaction with the volumes of the grapes (\eg harvesting), segmentation, and in particular instance segmentation, is required. Instance segmentation requires labels that are ideally pixel perfect masks, however Bellocchio et al. \citep{BellocchioRAL2019weakly} showed how, even with minimal labelling signal (e.g. presence or absence of an object in a image), the task network is able to learn representations that are close to masks of the object of interest. For this reason, we again adopt a pseudo-labelling approach to this problem, starting with a pretrained network on the WGISD source dataset and then using simple external cues to work as our external information signal that helps in refining the label. The overview of this sub-system is depicted in Figure \ref{fig:lower_subsys}.

\begin{figure}[t]
\centering
\includegraphics[width=0.5\textwidth]{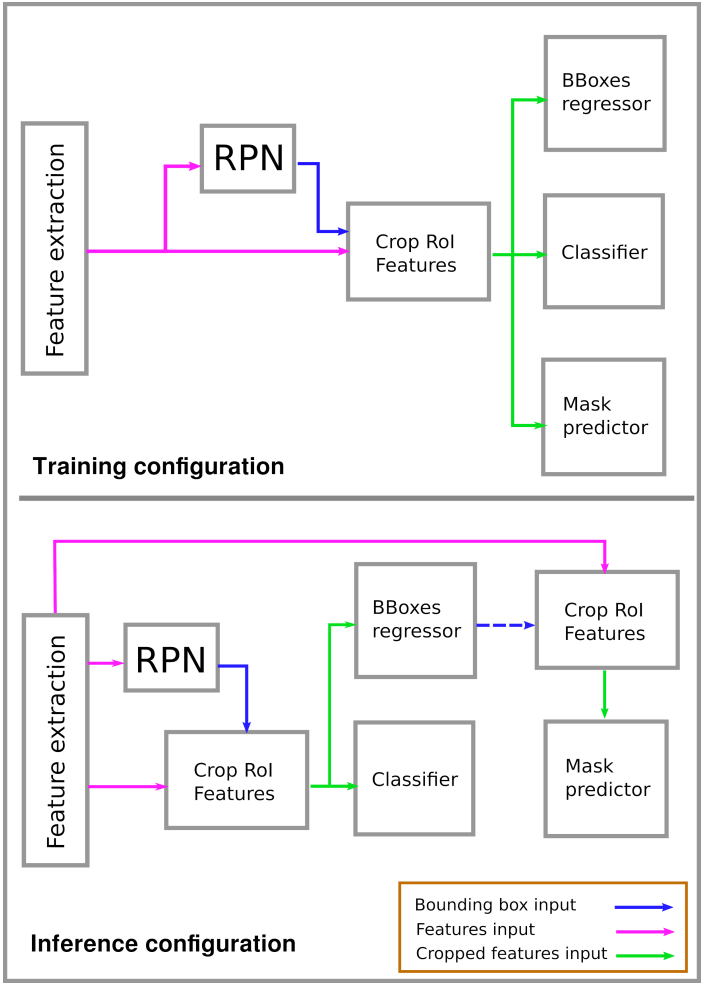}
\caption{Mask-RCNN internal wiring at training and inference times. At training time, the mask prediction head uses the same inputs as the other two heads, i.e. the RoI cropped features. At inference time, the cropping is done only using the bounding boxes proposals of the bounding box prediction head. Our system uses only the feature extraction part and drops the bounding box regression, using instead either the bounding boxes coming from ground truth or the bounding boxes pseudo-labels generated by the DPLG sub-system. The dashed blue line in the lower diagram shows where the wire is cut and our bounding boxes proposals are injected.}
\label{fig:maskrcnn_arch}
\end{figure}

\begin{figure}[h]
    \centering
    \begin{subfigure}[h]{0.4\columnwidth}
        \centering
        \includegraphics[]{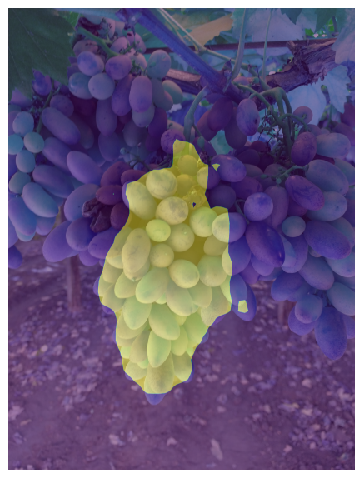}
    \end{subfigure}
    \begin{subfigure}[h]{0.4\columnwidth}
        \centering
        \includegraphics[]{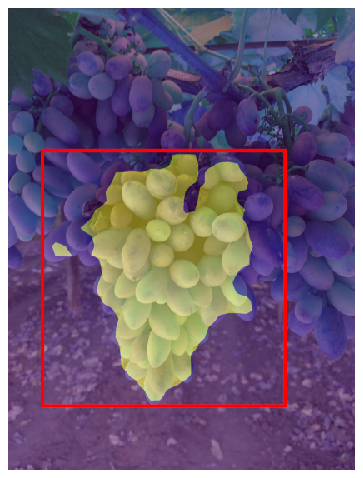}
    \end{subfigure}
    \caption{Left image: pseudo mask produced by Mask-RCNN trained only on the Source Dataset and without a bounding box cue. Right image: same image showing the effect of giving a bounding box cue at inference time.}
    \label{fig:attention_mechanism}
\end{figure}

Our SSeg network is Mask R-CNN trained on WGISD, as usual. Mask R-CNN in its basic form extracts region proposals and uses them to predict bounding boxes and instance segmentation masks. However, it 
is possible to use the segmentation subnetwork of Mask R-CNN as the pseudo mask initial generator. In particular, Mask R-CNN is wired differently at inference time than at training time, since the bounding boxes predicted by the detection head are directly fed to the mask head. The network will use this bounding box as a cue, or as an attention mechanism, which helps the segmentation sub network to output a useful pseudo mask. 
This is depicted in Figure \ref{fig:maskrcnn_arch}. 
This strategy will mitigate the problem of confirmation bias, since the box comes from an external information source. In our system, the bounding boxes could come from the output of the DPLG  sub-system. At the same time, in Section \ref{sec:experiments} we show the performances of the pseudo mask generation starting from ground truth bounding boxes so as to better isolate the performance contribution of the mask generation process. In this way, the number of pseudo-masks will coincide with the actual number of grape clusters in the image, and the measured error will only be due to the mask generation process. The qualitative difference of segmentation mask between the standard wiring of the Mask R-CNN network, and the one with an external attention mechanism, is shown in Figure \ref{fig:attention_mechanism}. 

\subsubsection{Pseudo Mask Refining Block} \label{sec:refining}
To refine the pseudo masks, in order to reduce or remove confirmation bias, an external source of information is needed. Some earlier works worked on this aspect, such as \citep{KhorevaCVPR2017simple}. We tried three different strategies to refine the initial masks, using simple computer vision techniques that work on different principles from the convolutional filters contained in SSeg and that use simple geometrical considerations.

\begin{figure*}[h]
    \centering
    \begin{subfigure}[h]{0.3\textwidth}
        \centering
        \includegraphics[width=\textwidth]{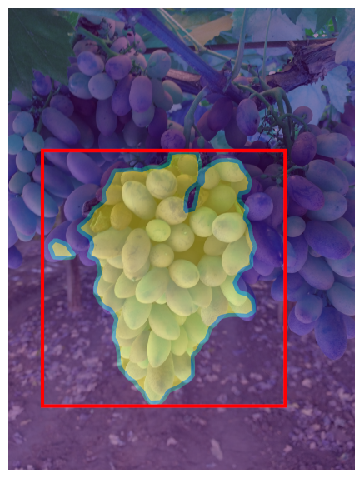}
        \caption{Dilation}
        \label{fig:dilation_refinement}
    \end{subfigure}
    \begin{subfigure}[h]{0.3\textwidth}
        \centering
        \includegraphics[width=\textwidth]{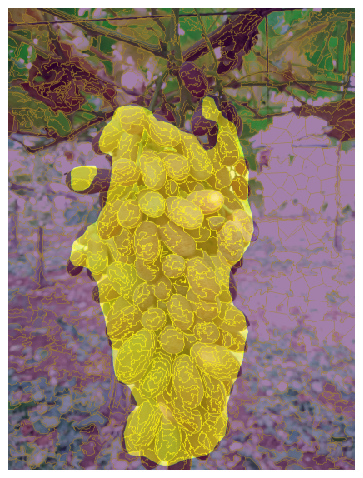}
        \caption{SLIC}
        \label{fig:slic_refinement}
    \end{subfigure}
    \caption{a) Example of application of the dilation method to the mask of a grape bunch. The yellow area represents the starting pseudo mask, while the green area represents the expansion done by the dilation operation. b) Example of the super-pixel segmentation (clustering) done by SLIC.}
\end{figure*}

\begin{figure*}[h]
    \centering
    \begin{subfigure}[h]{0.3\textwidth}
        \centering
        \includegraphics[width=\textwidth]{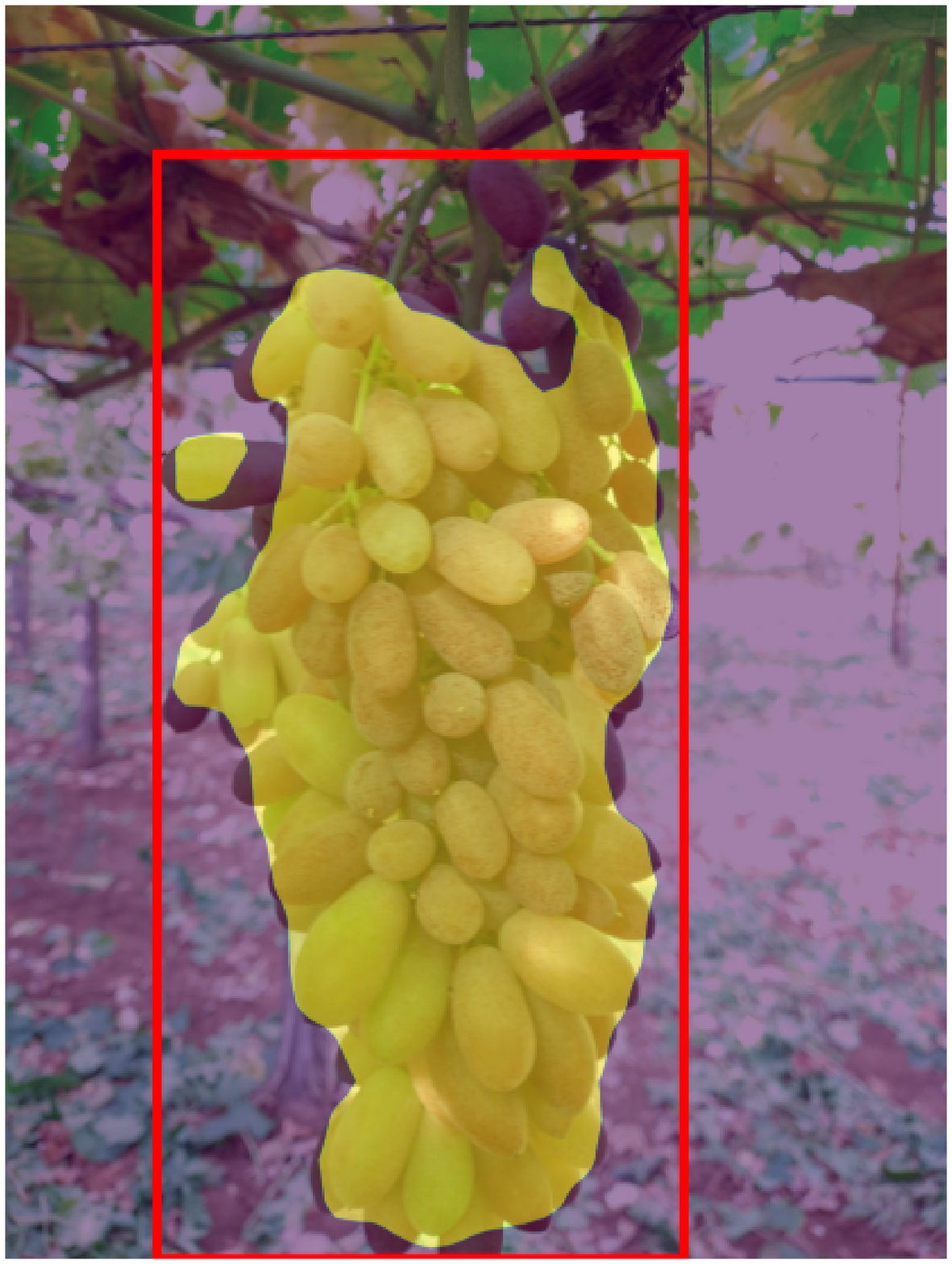}
    \end{subfigure}
    \begin{subfigure}[h]{0.3\textwidth}
        \centering
        \includegraphics[width=\textwidth]{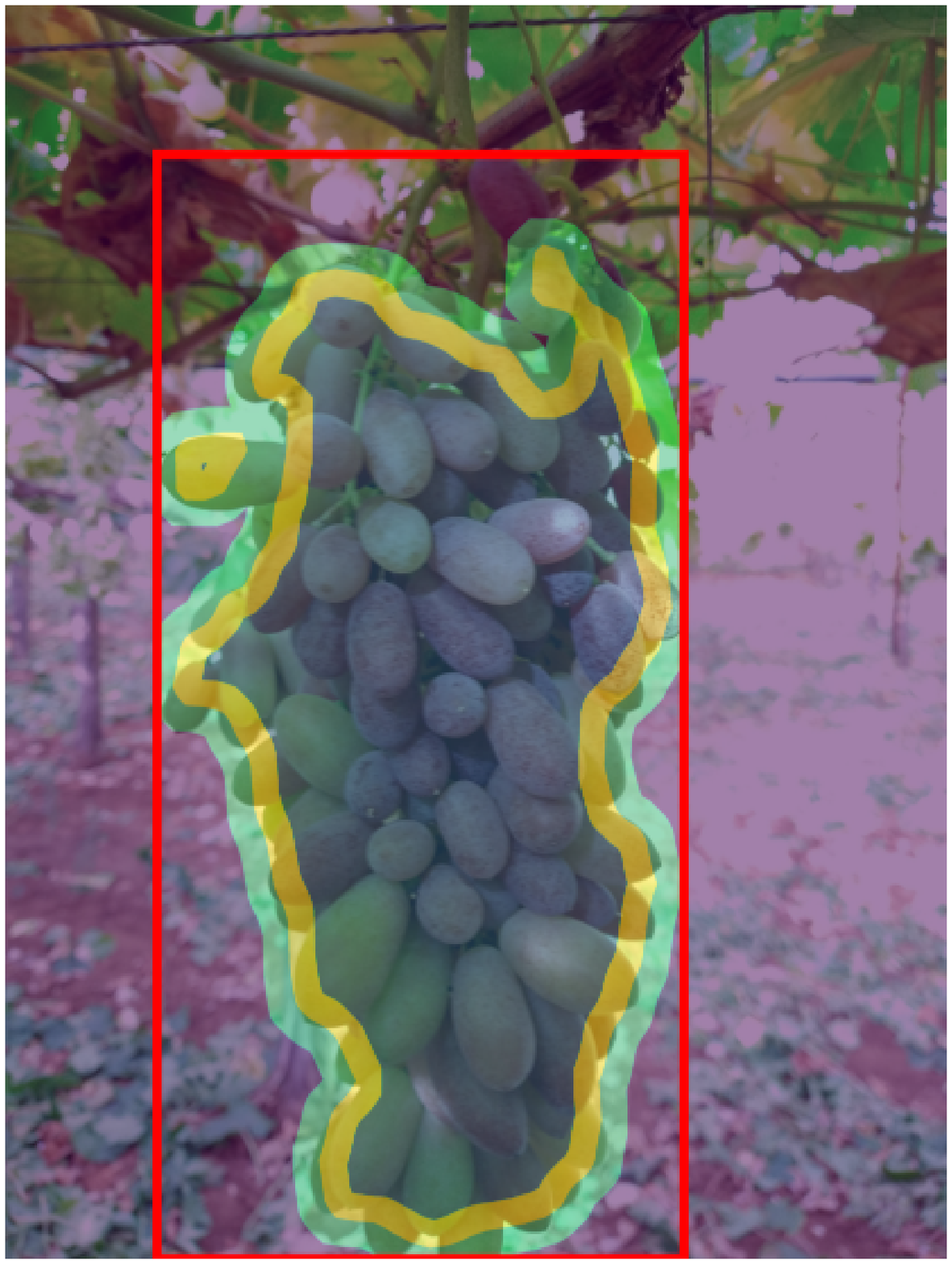}
    \end{subfigure}
    \begin{subfigure}[h]{0.3\textwidth}
        \centering
        \includegraphics[width=\textwidth]{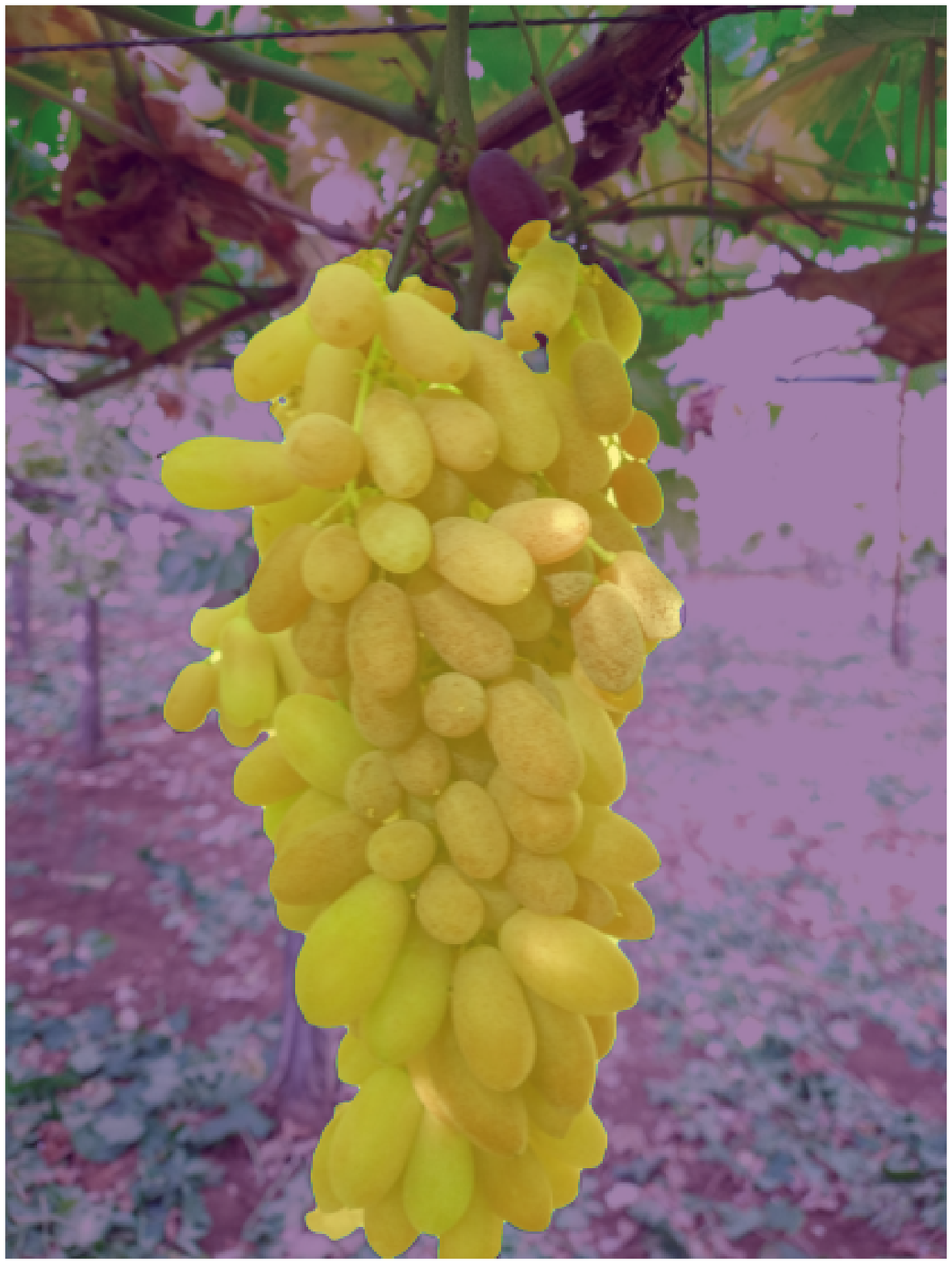}
    \end{subfigure}
    \caption{Examples of images refined with Grabcut. The color of the overlay defines the pixel as sure foreground (blue), probable foreground (yellow), probable background (green) or sure background (purple).}
    \label{fig:grabcut_refinement}
\end{figure*}

\begin{itemize}
    \item \textbf{Dilation}: the first method originates from the observation that SSeg tends to underestimate the masks on the target data. For this reason, a simple morphological dilation that expands the mask until it touches the reference bounding box is able to add valuable information to the label. The dilation is applied with a 5x5 circular-shaped kernel. An example of the result is given in Figure \ref{fig:dilation_refinement}.
    \item \textbf{SLIC}: Simple Linear Iterative Clustering (SLIC) \citep{Radhakrishna2012slic} is a method for super-pixel segmentation of the image. Super-pixels are contiguous regions of the image that are clustered together by a KMeans algorithm running on both color and space (5-dimensional). We apply this super-pixel division to the entire image and compare it with each pseudo mask. The SLIC algorithm that was used was the one implemented in the Python scikit-image library \citep{scikit-image} with 2000 segments and compactness 0.1. All the super pixels that are covered by more than an upper threshold $t_u = 70\%$ are added to the mask, while all the pixels that are covered by less than a lower threshold $t_l = 30\%$ are removed from the mask. The rationale is that in this way we should be able to remove also the background pixels erroneously contained in the initial pseudo mask. An example of the result is given in Figure \ref{fig:slic_refinement}.
    \item \textbf{Grub Cut}: this is an iterative segmentation technique introduced by \citep{rother2004grabcut}. It represents the image as a graph where foreground and background pixels are modeled as Gaussian Mixture Models and have to be separated iteratively by cuts to the graph edges. We used the OpenCV \citep{opencv_library}  implementation where it is possible to initialize the algorithm with the pseudo mask defining four pixel categories, \ie sure foreground, sure background, probable foreground, and probable background. The pseudo mask is used as probable foreground. Dilation is applied to the pseudo mask for a number of iterations proportional to the smallest dimension of the reference bounding box to obtain the probable background. Erosion is applied for the same number of iterations to obtain the sure foreground, while the rest is set to sure background. A sample of the effects of Grab Cut is shown in Figure  \ref{fig:grabcut_refinement}.
\end{itemize}

In Section \ref{sec:experiments}, we show how each of these refinement methods performs compared to the baseline (pseudo mask with no refinement). 

\section{Results and Discussion} \label{sec:experiments}

\subsection{Detection Experiments} \label{sec:detection_results}
In this Section, we describe the results of the detection experiments. Table \ref{tab:yolov5models} shows our preliminary experiments to compare different versions of the detector. All models in this initial comparison are trained and tested on WGISD. Results show that the models with a large number of parameters offer a minimal performance increase on basic detection, compared to the lightweight versions S and N. This can be explained by a general homogeneity of the distribution of the grape images in WGISD, which does not require huge number of parameters to learn a good estimator. This is expected, since in agriculture we do not work with huge amounts of data. For this reason, we decided to base all the trackers on the S and N variants to reduce overfitting.

\begin{table}[!h]
\caption{\label{tab:yolov5models} Comparison of the YOLOv5 models tested to be the tracker engine, trained and tested on WGISD. The models with the highest number of parameters do not have a significant performance advantage over the lightweight versions S and N.}
\centering
\resizebox{\columnwidth}{!}{%
    \begin{tabular}{|c|c|c|c|c|}
    \hline
        Model & $mAP_{0.5:0.95}$ & $mAP_{0.5}$ & Speed (ms) & Params (M)\\ \hline \hline
        YOLOv5n & 58.2 & 89.4 & 6.3 & 1.9\\ \hline
        YOLOv5s & 62.5 & 89.7 & 6.4 & 7.2\\ \hline
        YOLOv5m & 61.9 & 89.5 & 8.2 & 21.2\\ \hline
        YOLOv5l & 64.0 & 90.5 & 10.1 & 46.5\\ \hline
        YOLOv5x & 61.5 & 87.5 & 12.1 & 86.7\\ \hline
    \end{tabular}
}
\end{table}

\subsubsection{Training Details} \label{sec:sdet}
All the training experiments conducted on YOLO have been done on the Nvidia DGX-1 Station, since it offers an appropriate computational power for the training. All the training runs are composed of 300 epochs with a batch size of 4 and the patience parameter for early stopping set to 30 epochs. The learning rate ($lr$) strategy used is \textit{one cycle} \citep{Smith2018superconvergence}, with initial $lr=0.01$ and final $lr=0.001$. The optimizer is SGD, with momentum 0.937 and weight decay \num{5e-4}. The time required for training was highly influenced by the specific version of YOLO and by the number of images used, meaning that using 3368 images for training, the smaller version of YOLO (YOLOn) required just under 3 hours for all the epochs, while the bigger one (YOLOx) took more than 24 hours to converge.
All the detection networks were pretrained on MS COCO dataset \citep{Lin2014microsoft} and then finetuned on the source and target datasets. The baseline model was trained only on the source dataset, while the proposed models were trained on the target using the methods described in \ref{sec:bbox_interpolation}. In order to help generalization to different conditions, the 242 training images of the source set were augmented with random crop, random contrast, Gaussian blur, Gaussian noise and horizontal flip. During our experiments, these random augmentations have been applied offline four times, generating 726 augmented images.

Table \ref{tab:SDetVsTDet} shows the difference on the images of the target set ($T_{img}$) between the detector trained only on the source data (SDet) and also on the pseudo-labels generated from the videos (TDet). It is possible to see that the $mAP_{0.5}$ increased by \textbf{8\%} despite the fact that the videos have a different distribution with respect to the images, due to the different process followed to collect them.

\begin{table}[!h]
\caption{\label{tab:SDetVsTDet} Comparison of the source detector (SDet) with the target one (TDet) that has been trained also using the pseudo-labels generated from the videos. The numbers shown are the performance on the test set of the TImg data.}
\centering
    \begin{tabular}{|c|c|c|c|c|}
    \hline
        \multicolumn{5}{|c|}{Detectors Performance on the TImg dataset}\\
    \hline
        Model & Precision & Recall & $mAP_{0.5}$ & $mAP_{0.95}$\\ \hline \hline
        SDet & 0.90 & 0.56 & 0.69 & 0.46\\ \hline
        TDet & 0.98 & 0.68 & 0.77 & 0.47\\ \hline
    \end{tabular}
\end{table}

\begin{figure*}[h]
\centering

	\begin{subfigure}[t]{0.2\textwidth}
		\includegraphics[width=\columnwidth]{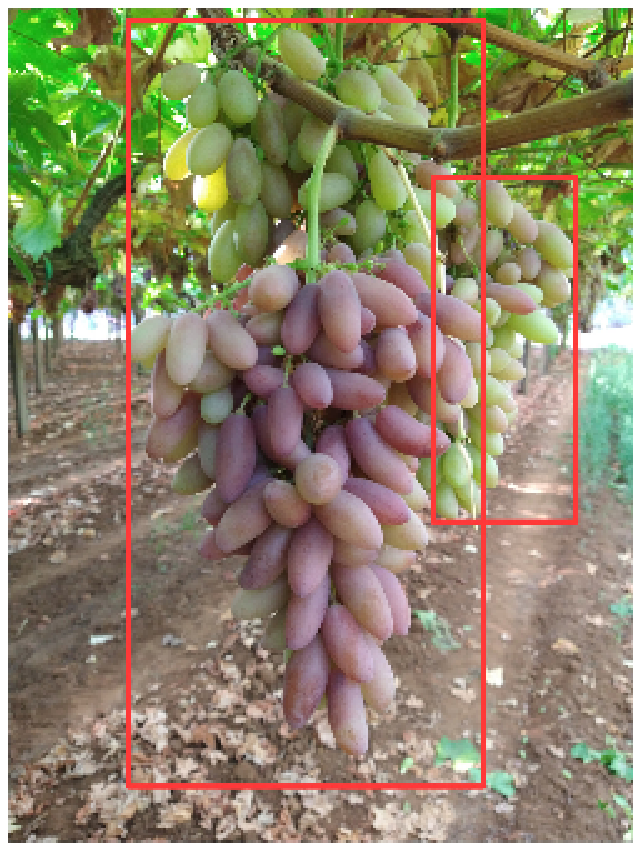}
		\caption{}
	\end{subfigure}
	\begin{subfigure}[t]{0.2\textwidth}
		\includegraphics[width=\columnwidth]{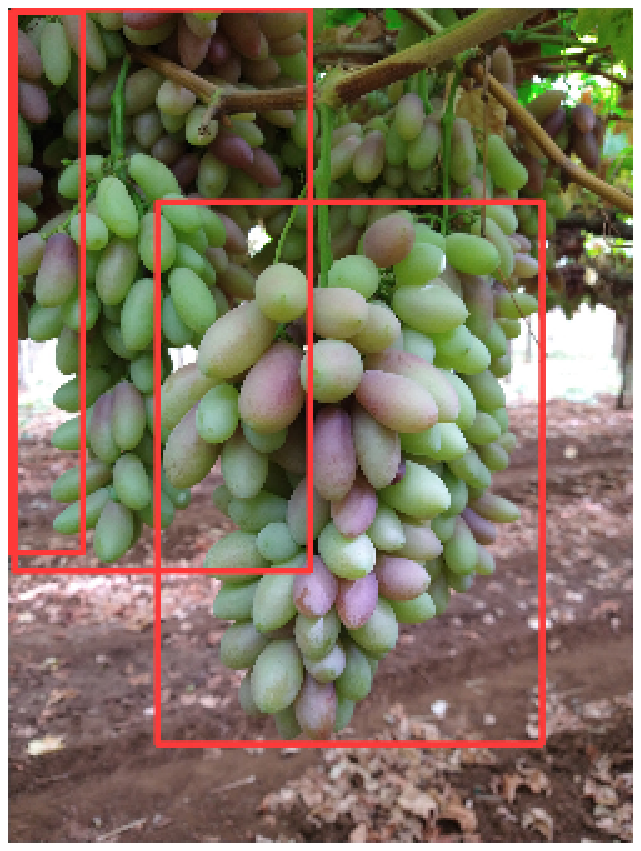}
		\caption{}
	\end{subfigure}
	\begin{subfigure}[t]{0.2\textwidth}
		\includegraphics[width=\columnwidth]{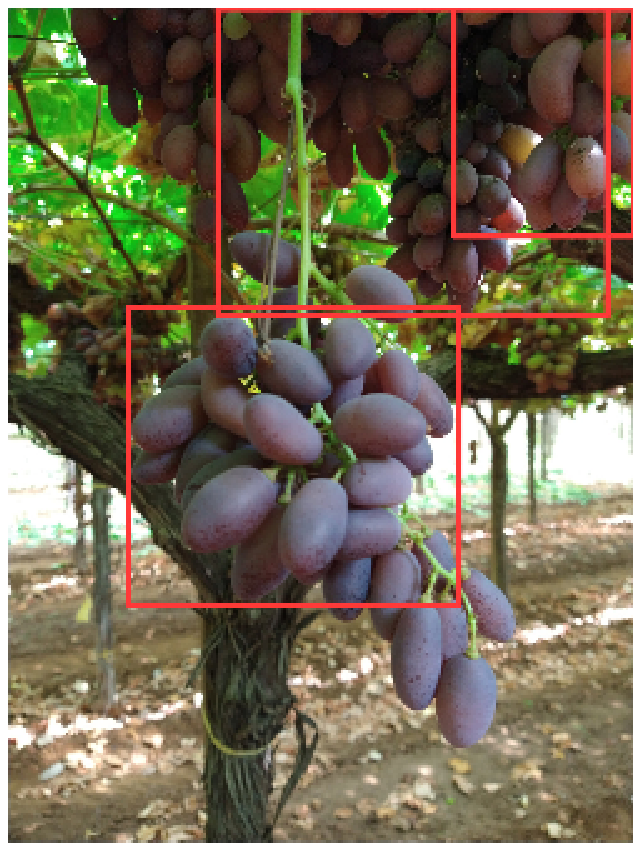}
		\caption{}
	\end{subfigure}
	\begin{subfigure}[t]{0.2\textwidth}
		\includegraphics[width=\columnwidth]{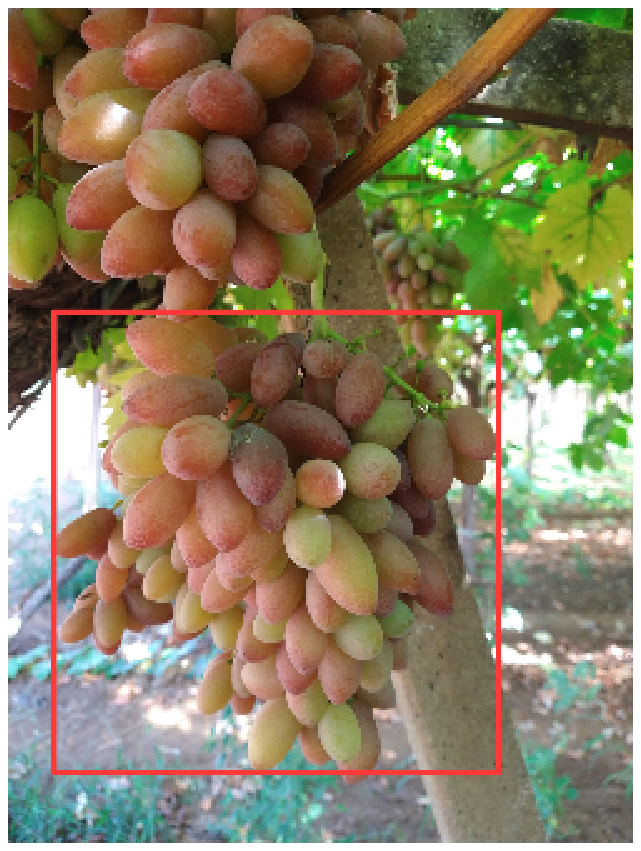}
		\caption{}
	\end{subfigure}\\
	\begin{subfigure}[t]{0.2\textwidth}
		\includegraphics[width=\columnwidth]{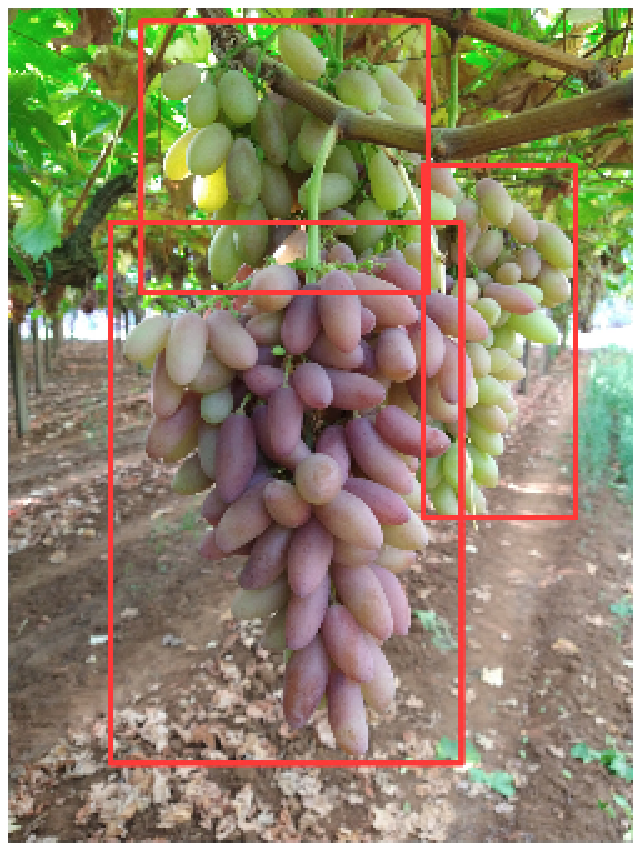}
		\caption{}
	\end{subfigure}
	\begin{subfigure}[t]{0.2\textwidth}
		\includegraphics[width=\columnwidth]{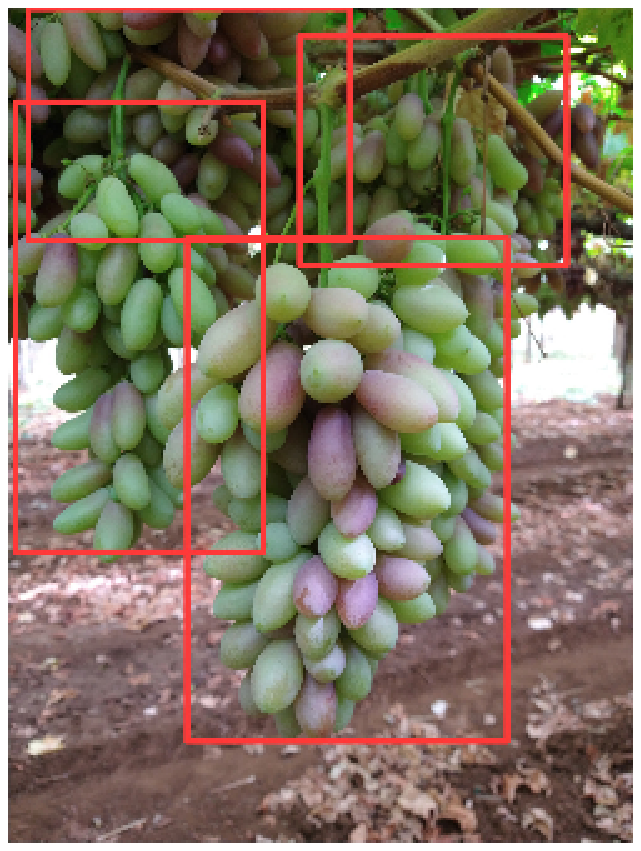}
		\caption{}
	\end{subfigure}
	\begin{subfigure}[t]{0.2\textwidth}
		\includegraphics[width=\columnwidth]{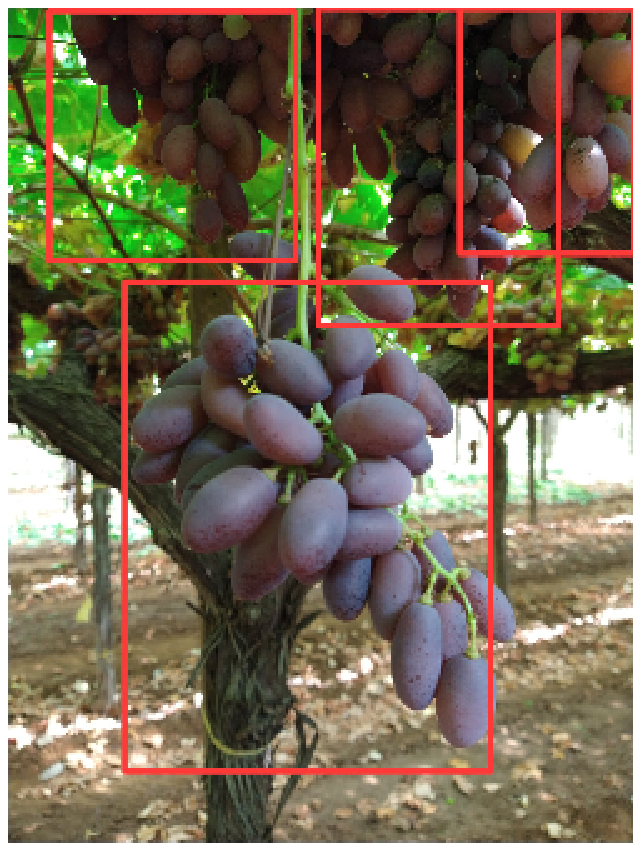}
		\caption{}
	\end{subfigure}
	\begin{subfigure}[t]{0.2\textwidth}
		\includegraphics[width=\columnwidth]{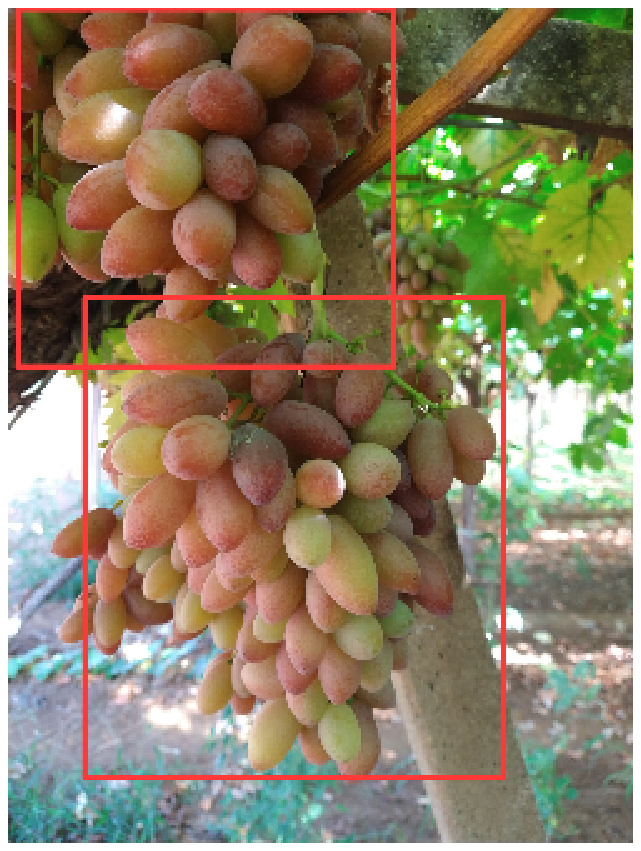}
		\caption{}
	\end{subfigure}

\caption{Examples of how the detection improves on the same images when the model is trained without the pseudo-labels (SDet) in the upper row (a,b,c,d), and with the pseudo-labels (TDet) in the lower one (e,f,g,h).}
\label{fig:detection_change}
\end{figure*}

An example of how the detection changes is given in Figure \ref{fig:detection_change}. In the upper row, are present the detections made by SDet, while in the lower one there are the predictions made by TDet. It is possible to see that not only the bounding boxes are tighter around the instances, but also more grapes are detected, meaning that both precision and recall have improved. 

Since TImg and TVid have different distribution, we also applied TDet on the test data from TVid. 
The results are shown in Table \ref{tab:SDetVsTDet_TVid}, where it is possible to see that the network trained with the pseudo-labels gained \textbf{10\%} in $mAP_{0.5}$ compared to the one trained without them. In this case, the increase is higher due to the minimal covariate shift between TVid test and training data. 

\begin{table}[!h]
\caption{\label{tab:SDetVsTDet_TVid} Comparison of the source detector (SDet) with the target one (TDet) that has been trained also using the pseudo-labels generated from the videos. The numbers shown are the performance on the test set of the TVid data.}

\centering
    \begin{tabular}{|c|c|c|c|c|}
    \hline
        \multicolumn{5}{|c|}{Detectors Performance on the TVid dataset}\\
    \hline
        Model & Precision & Recall & $mAP_{0.5}$ & $mAP_{0.95}$\\ \hline \hline
        SDet & 0.62 & 0.59 & 0.55 & 0.21\\ \hline
        TDet & 0.74 & 0.60 & 0.65 & 0.23\\ \hline
    \end{tabular}
\end{table}

\subsection{Tracking Experiments} \label{sec:tracking_results}
\begin{figure*}[h]
    \centering
    \includegraphics[width=\textwidth]{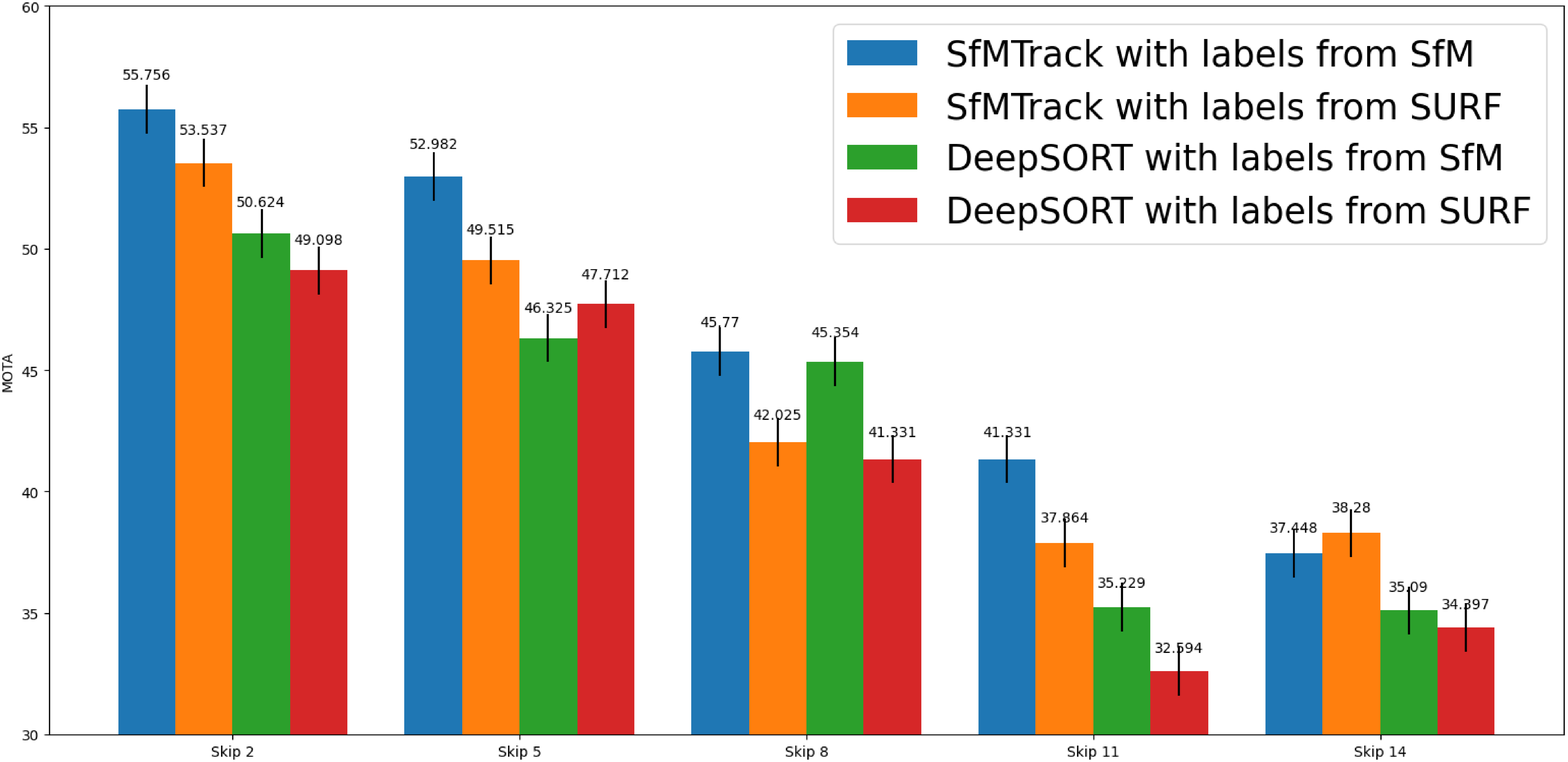}
    \caption{Comparison of trackers performances as the \textit{skip} value changes. The skip value is the number of frames skipped between a keyframe and the next in the process of generating pseudo-labels. The extracted pseudo-labels influence the detector performance, both due to quantity and quality of the labels, and consequently also the tracker is influenced. In this chart, we show the degradation of performances as the hyper-parameter is increased. While the best performances are obtained on skip 2, the degradation with skip 5 could be tolerable considering that it requires only 20\% of labelled frames instead of 50\%.}
    \label{fig:skip_cv}
\end{figure*}

In this Section, we describe the results of the tracking experiments. Each tracker is built on a YOLOv5 detector version. As explained in Section \ref{sec:tracking}, we compare two tracking schemes combined with two pseudo label generation strategies. Pseudo-labels depend, among other intrinsic and extrinsic variables, on the frame rate combined with the \textit{skip} value, as was described in Section \ref{sec:bbox_interpolation}. As mentioned in that section, with the bounding box interpolation system we use the best performing skip value which is 2, as shown in Figure \ref{fig:skip_cv}. This is to be expected, because if too many frames are interpolated, the motion becomes too large to be compensated. In addition, from the same results it is clear that the SfM approach has higher MOTA than DeepSORT at most skip values due to its use of the geometrical representation of the scene. However, it is not meant for real time computation. 
 In Table \ref{tab:sfm_deepsort_comparison1}, the MOT metrics for the best models have been summarized.

\begin{figure*}[!h]
    \centering
    \begin{subfigure}[t]{0.3\textwidth}
        \centering
        \includegraphics[width=\textwidth]{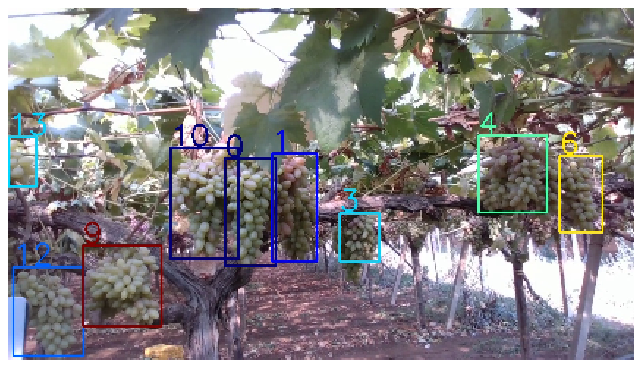}
        \caption{I1}
    \end{subfigure}
    \begin{subfigure}[t]{0.3\textwidth}
        \centering
        \includegraphics[width=\textwidth]{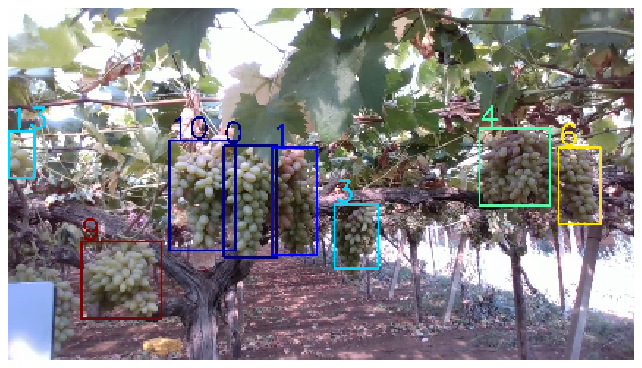}
        \caption{I2}
    \end{subfigure}
    \begin{subfigure}[t]{0.3\textwidth}
        \centering
        \includegraphics[width=\textwidth]{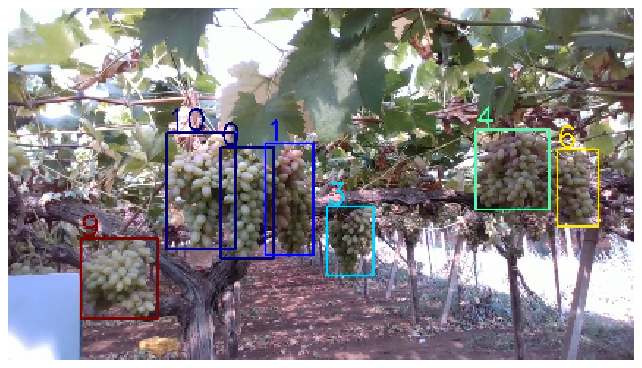}
        \caption{I3}
    \end{subfigure}\\
    \begin{subfigure}[t]{0.3\textwidth}
        \centering
        \includegraphics[width=\textwidth]{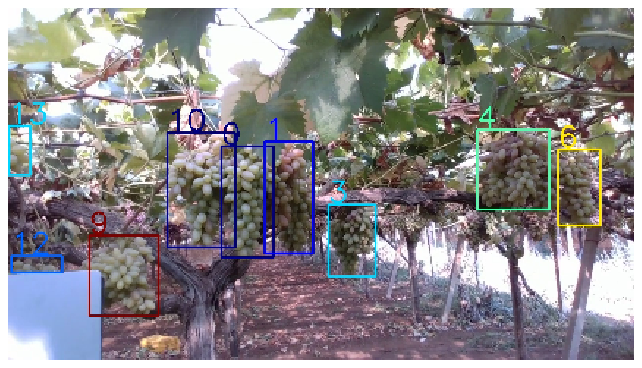}
        \caption{I4}
    \end{subfigure}
    \begin{subfigure}[t]{0.3\textwidth}
        \centering
        \includegraphics[width=\textwidth]{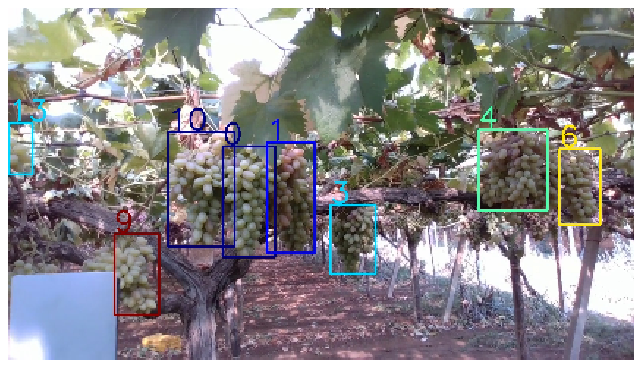}
        \caption{I5}
    \end{subfigure}
    \begin{subfigure}[t]{0.3\textwidth}
        \centering
        \includegraphics[width=\textwidth]{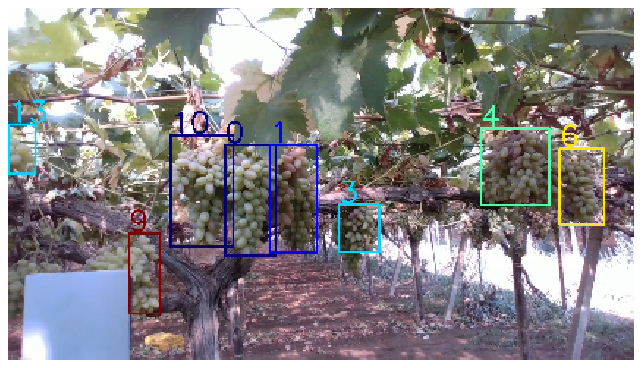}
        \caption{I6}
    \end{subfigure}\\
    \begin{subfigure}[t]{0.3\textwidth}
        \centering
        \includegraphics[width=\textwidth]{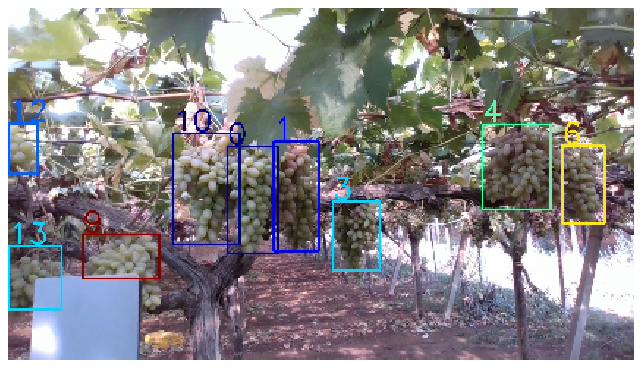}
        \caption{I7}
    \end{subfigure}
    \begin{subfigure}[t]{0.3\textwidth}
        \centering
        \includegraphics[width=\textwidth]{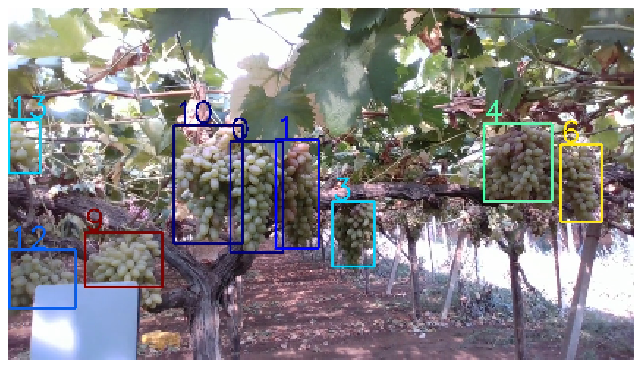}
        \caption{I8}
    \end{subfigure}
    \begin{subfigure}[t]{0.3\textwidth}
        \centering
        \includegraphics[width=\textwidth]{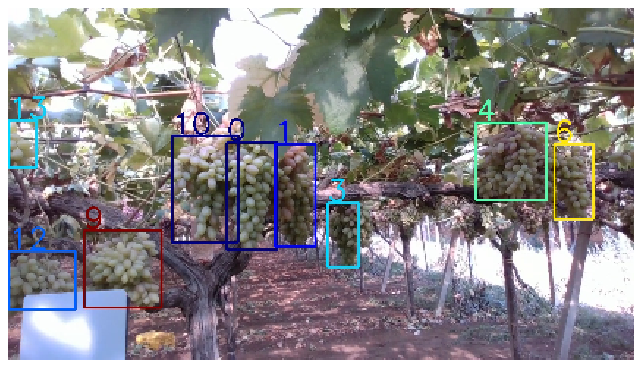}
        \caption{I9}
    \end{subfigure}
	\begin{subfigure}[b]{0.8\textwidth}
        \centering
        \includegraphics[width=\textwidth]{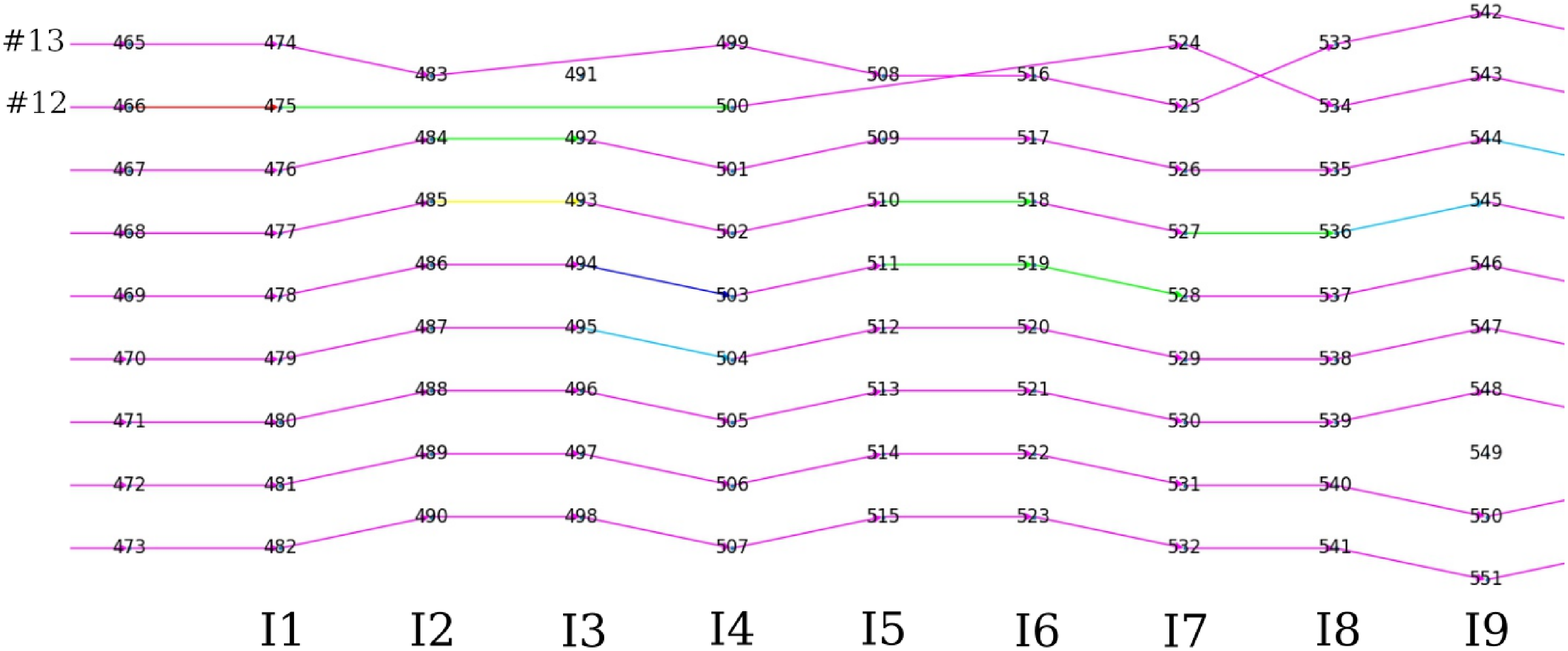}
        \caption{Tracking diagram: the two uppermost tracks relate to the \#12 and \#13 IDs.}
    \end{subfigure}
    \caption{Examples of failure cases in SfM tracking: The sequence of frames shows some examples of fragmented trajectories and ID switches. It can be seen how ID\#12 is occluded and disappears in frames I2 and I3, but it is correctly associated in frame I4. It is occluded again in frames I5 and I6, and erroneously switched with ID\#13 in frame I7, but recovered in frame I8. These count for two ID switches, while the total number of tracks remains the same, so the final yield count is not affected by these errors. }
    \label{fig:id_switch}
\end{figure*}

Among the MOT metrics described in Section \ref{sec:metrics}, the MOTA and MOTP give a general idea of the tracking performances. However, to use trackers as yield estimators, one of the figures of interest is the number of IDs that the tracker finds, which could be considered an estimate of the number of bunches found. However, during the tracking process, some of the bunch IDs are switched. This can happen, for example, when two bunches are occluding each other and the IDs are inverted after the occlusion situation disappears, or when there are errors in feature association in the SfM block (Figure \ref{fig:id_switch}). Whatever the reason, this situation is captured by the $ID_{sw}$ metric, together with the MOTA score. More details on these challenges of MOT can be found in \citep{clearmot}. For all these reasons, we focus our attention on MOTA for the tracking accuracy, and on the number of IDs for the yield estimation.

\begin{table*}[th]
\caption{\label{tab:sfm_deepsort_comparison1}The upper half of this table shows some performance metrics for two types of trackers. Both trackers are based on a YOLOs detector, trained with two different datasets: \textbf{WGISD} is the baseline, while the \textbf{Pseudo-labels} is trained on pseudo-labels generated as described in Section \ref{sec:bbox_interpolation}.
The lower half shows other common MOT metrics, notably $IDs$ is the one used to compute yield estimation. It can be seen that for both tracking strategies there is a consistent advantage in using the pseudo-labels. In particular, the error in yield estimation for the SfM tracker drops by 29\%.}

\centering
	\resizebox{0.8\textwidth}{!}{%
    \begin{tabular}{|c|c|c|c|c|c|c|c|c|} \hline
        Method & MOTA$\uparrow$ & MOTP$\uparrow$ & MT$\uparrow$ & ML$\downarrow$ & $ID_{sw}\downarrow$ & FM$\downarrow$ & Pr$\uparrow$ & Re$\uparrow$\\ \hline
        SfMTrack \textbf{WGISD} & 46.741 & 72.545 & 9 & 9 & 5 & 29 & 91.304 & 52.427\\
        SfMTrack \textbf{Pseudo-labels} & 55.756 & 74.557 & 11 & 8 & 9 & 22 & 89.143 & 64.91\\
        \hline
        DeepSort \textbf{WGISD} & 40.499 & 72.229 & 8 & 8 & 16 & 19 & 87.198 & 50.069\\
        DeepSort \textbf{Pseudo-labels} & 50.624 & 72.941 & 9 & 6 & 17 & 18 & 85.634 & 63.662\\
        \hline
        \hline
        Method & TP$\uparrow$ & FP$\downarrow$ & FN$\downarrow$ & Dets & GT Dets & IDs & GT IDs & Yield est. Err\\ \hline
        SfMTrack \textbf{WGISD} & 344 & 70 & 377 & 414 & 721 & 19 & 31 & 38\%\\
        SfMTrack \textbf{Pseudo-labels} &  431 & 94 & 290 & 525 & 721 & 28 & 31 & \textbf{9\%} \\
        \hline
        DeepSort \textbf{WGISD} &  299 & 115 & 422 & 414 & 721 & 46 & 31 & 48\%\\
        DeepSort \textbf{Pseudo-labels} & 380 & 156 & 341 & 536 & 721 & 39 & 31& \textbf{26\%} \\
        \hline
    \end{tabular}
    }
\end{table*}

From Table \ref{tab:sfm_deepsort_comparison1} it is evident that the pseudo label generation is highly beneficial, with an increase of more than \textbf{10\%} compared to the source dataset WGISD. Looking closely at the performance metrics, it can be seen that the capacity to track the same IDs for the entire trajectory is stronger in DeepSORT. This is probably due to Kalman filtering, since taking into account the bounding box movement dynamics implicitly avoids errors such as the one shown in Figure \ref{fig:id_switch}.

\subsection{Segmentation Experiments} \label{sec:segmentation_results}
In this Section, we show the results of the experiments concerning the performance of the SPLG sub-system. The pseudo-mask generator can be seen either an independent system or in conjunction with the DPLG sub-system. In the following, the experiments of the SPLG sub-system are described, while the results of the whole system are described in Section \ref{sec:joint_sys_results}.

\subsubsection{Training Details} \label{sec:sseg}
The implementation of Mask R-CNN we chose is Detectron2 \citep{Wu2019detectron2}, using ResNet 101 as backbone network. Again, the experiments were performed on the NVidia DGX cluster. The training started from the MS COCO weights, then was fine-tuned on the source and target dataset. For all the training, common data augmentation was performed by applying Gaussian blur, Gaussian noise, random changes in brightness and contrast, pixel dropout, random flip, and random crop. In addition, the trainings were executed using a learning rate of 0.001, weight decay of 0.0001 and a momentum of 0.9. Each training proceeded for a maximum of 100 epochs, but early stopping was used while monitoring the segmentation AP on the validation set of the table grape dataset, with a patience value of 20.

As for the detector, we give an idea of the initial performance gap of SSeg when directly applied to the target dataset TImg  in Table~\ref{tab:coco_wgisd}, using the MS COCO \citep{Lin2014microsoft} metrics as described in Section \ref{sec:metrics}. We performed data augmentation on the WGISD dataset, in particular, crop and resize to mitigate the difference in scale with the TD. Nonetheless in all metrics there are more than 20 points of decrement in Average Precision.

\begin{table}[h]
\caption{\small Evaluation of covariate shift for SSeg: SSeg is a Mask R-CNN model trained on the Source dataset (WGISD) and in this table we show the performance comparison when it is tested on WGISD test set (27 images) and on the test set of our TImg dataset (20 images) using some of the COCO metrics.}
    \centering
    \resizebox{\columnwidth}{!}{%
    \begin{tabular}{|c||c|c|c|c|}
    \hline
        Test data & Task & $mAP$ & $mAP_{0.5}$ & $mAP_{0.75}$ \\
    \hline
        \multirow{2}{*}{WGISD} & Detection & 53.40 & 87.02 & 57.36 \\ 
    \cline{2-5}
        & Segmentation & 53.60 & 89.44 & 55.41 \\
    \hline
        \multirow{2}{*}{TImg} & Detection & 32.65 & 60.40 & 30.37 \\
    \cline{2-5}
        & Segmentation & 32.88 & 65.40 & 34.77 \\
    \hline
    \end{tabular}
    }
    
    \label{tab:coco_wgisd}
\end{table}

\subsubsection{Pseudo-mask generation experiments}

\begin{table*}[!h]
\caption{\label{tab:refining_comparison} Comparison of the Mask R-CNN model trained with different pseudo-mask processing methods.}
\centering
    \begin{tabular}{|c|c|c|c|}
    \hline
        Training Data & $mAP_{0.5:0.95}$ & $mAP_{0.5}$ & $mAP_{0.75}$ \\ \hline \hline
        WGISD (baseline) & 32.88 & 65.40 & 34.77 \\ \hline
        WGISD + TImg & 48.43 & 83.06 & 53.12\\ \hline
        WGISD + TImg w/ Dilation & 48.67 & 81.54 & 54.87 \\ \hline
        WGISD + TImg w/ SLIC & 47.78 & 80.41 & 53.54 \\ \hline
        WGISD + TImg w/ Grabcut & 49.56 & 81.03 & 57.70 \\ \hline
    \end{tabular}
\end{table*}

The first set of experiments are an ablation study to evaluate the performance of TSeg in isolation from TDet in order to quantify the effectiveness of generating pseudo-labels when no other mask labels on target data are provided. As before, SSeg is our baseline and in this case TSeg is trained on both the source dataset and the training set of TImg, whose labels were generated as pseudo masks by SSeg with the successive refinement.

We performed comparison experiments between the three refinement strategies presented in Section \ref{sec:refining}, namely dilation, SLIC and GrabCut. We show in Table~\ref{tab:refining_comparison} the average performance of five trials for each experiment, as evaluated on the TImg test set. In the same table, we show the results obtained by TSeg trained with and without the Refining Block. The additional pseudo masks are able to considerably improve the performance on the target data in terms of AP, with an improvement of almost \textbf{50}\% on the baseline performance. Moreover, our results show that the best performing refining method is GrabCut. The additional refinement increases the $mAP_{0.5:0.95}$ by $\mathbf{1.13}$ and the $mAP_{0.75}$ by $\mathbf{4.58}$ with respect to TSeg trained without refinement, but decreases in $mAP_{0.5}$, showing that the refinement process is more effective at higher IoU levels.

\subsection{Complete System Experiments} \label{sec:joint_sys_results}
In this Section, we describe the results obtained by using the detector described in section \ref{sec:detection_results} to generate the bounding boxes required by the SPLG sub-system described in the previous section \ref{sec:segmentation_results}. 
First the best YOLOv5 detector, namely that obtained with the use of the pseudo-labels generation method, was used to predict the bounding boxes that are used to generate the pseudo-masks by Mask R-CNN. This was done both for TImg training set and for TVid. 
The test data for this experiment is the TImg test set, so the training and test distributions, although they are target data, are different. 
Table \ref{tab:semi_supervised_comparison} again shows the comparison of TSeg with and without the Refining Block. In the case of refined TImg, the improvement is still substantial, more than \textbf{40}\% over the baseline. Moreover, TVid is able to give an even greater improvement thanks to the greater number of images in the training set. Despite the fact that the video frames present many differences with respect to the target dataset, the TSeg still manages to increase the performance by \textbf{42\%} with respect to the baseline. Finally, the increase in mAP is even greater when considering TImg and TVid as training data. Also in those experiments, the Refinement Block gives an improvement over the non refined counterpart. From the values of $mAP_{0.50}$ and $mAP_{0.75}$ we deduce that the increase is mainly due to an improvement at IoU higher than $0.75$.

\begin{table*}[!h]
\caption{\label{tab:semi_supervised_comparison} Comparison of the Mask R-CNN model trained on different training sets with bounding boxes generated by YOLO (TDet). The number of images in the training set are shown in parenthesis.}
\centering
    \begin{tabular}{|c|c|c|c|}
    \hline
        Training Data & $mAP_{0.5:0.95}$ & $mAP_{0.5}$ & $mAP_{0.75}$ \\ \hline \hline
        WGISD (baseline) (88) & 32.88 & 65.40 & 34.77 \\ \hline
        WGISD + TImg (182) & 44.45 & 75.68 & 49.78 \\ \hline
        WGISD + TImg (182) w/ Grabcut & 46.41 & 78.09 & 51.74 \\ \hline
        WGISD + TVid (687) & 45.99 & 78.30 & 52.59 \\ \hline
        WGISD + TVid (687) w/ Grabcut & 46.66 & 77.38 & 52.22 \\ \hline
        WGISD + TImg + TVid (781) & 47.44 & 76.63 & 56.06 \\ \hline
        WGISD + TImg + TVid (781) w/ Grabcut & 47.81 & 77.23 & 53.27 \\ \hline
    \end{tabular}
\end{table*}

\section{Conclusions} \label{sec:conclusions}

In this work, a system for producing pseudo-labels for detection and segmentation tasks has been presented. This system is particularly aimed at agricultural applications, where data scarcity is a common challenge. 
The system has two components, the Detection Pseudo-Label Generator and the Segmentation Pseudo-Label Generator. Both sub-systems require a starting coarse detection, or segmentation learning algorithm, respectively, to find the initial labels estimates. This is not a difficult requirement to fulfil, since the initial performances do not have to be high, and a limited amount of data, even from a different dataset, have have been shown to be sufficient.  
The detection PLG is able to label any data collected from simple continuos videos of the target objects by leveraging the 3D structure extracted from the video motion.
The segmentation PLG is able to work on any image, and uses other segmentation strategies to refine the pseudo-labels produced by the initial segmentation algorithm. The two subsystems can be chained in a single PLG system able to extract both bounding boxes and segmentation masks from the video provided.  
New detection and segmentation algorithms can be trained on the pseudo-labels and the experiments show that their performances surpass the initial algorithms performances by a large margin.

Although demonstrated on the problem of table grape labelling with covariate shift, the system can be applied to other fruits. 
This approach could be used also as part of more sophisticated and expensive agronomic solutions, such as robotic harvesting systems, leading to savings in the labelling costs and in development time. Future development will address iterative pseudo-label refinement and the removal of initial requirements to make the system fully unsupervised. 

\subsection*{CRediT authorship contribution statement}
\textbf{Thomas A. Ciarfuglia}: Conceptualization, Methodology, Validation, Investigation, Data Curation, Writing - Original Draft, Writing - Review \& Editing, Visualization, Supervision, Project administration;
\textbf{Ionut M. Motoi}: Methodology, Software, Validation, Formal analysis, Investigation, Data Curation, Writing - Original Draft;
\textbf{Leonardo Saraceni}: Methodology, Software, Validation, Formal analysis, Investigation, Data Curation, Writing - Original Draft;
\textbf{Mulham Fawakherji}: Methodology; 
\textbf{Alberto Sanfeliu}: Methodology, Supervision;
\textbf{Daniele Nardi}: Supervision, Project administration, Resources, Funding acquisition, Writing - Review \& Editing;

\section*{Acknowledgment}
\euflag \quad This work has been supported by the European Commission under the grant agreement number 101016906 – Project CANOPIES


\bibliographystyle{elsarticle-harv}
\bibliography{bibliography}
\end{document}